\documentclass[journal]{IEEEtran}
\usepackage{amsthm,amssymb,url,amsmath}
\usepackage{graphicx}
\usepackage{subfigure}
\usepackage{multicol}

\usepackage{cite}

\ifCLASSINFOpdf
\else
\fi
\usepackage{url}

\begin{document}
%
\title{Space Adaptive Search for Nonholonomic Mobile Robots Path Planning}
%
%
%

\author{Qi Wang
    \thanks{Qi Wang is with Autonomous Driving Lab, Tencent, China.\newline
    {\tt\small wangqi.dream@gmail.com} }%
}

\maketitle

\begin{abstract}
  Path planning for a nonholonomic mobile robot is a challenging problem. This paper proposes a novel space adaptive search (SAS) approach that greatly reduces the computation cost of nonholonomic mobile robot path planning. The classic search-based path planning only updates the state on the current location in each step, which is very inefficient, and, therefore, can easily be trapped by local minimum.
  The SAS updates not only the state of the current location, but also all states in the neighborhood, and the size of the neighborhood is adaptively varied based on the clearance around the current location at each step. Since a great deal of states can be immediately updated, the search can explore the local minimum and get rid of it very fast. As a result, the proposed approach can effectively deal with clustered environments with a large number of local minima.
  The SAS also utilizes a set of predefined motion primitives, and dynamically scales them into different sizes during the search to create various new primitives with differing sizes and curvatures. This greatly promotes the flexibility of the search of path planning in more complex environments. 
  Unlike the A* family, which uses heuristic to accelerate the search, the experiments shows that the SAS requires much less computation time and memory cost even without heuristic  than the weighted A* algorithm, while still preserving the optimality of the produced path. However, the SAS can also be applied together with heuristic or other path planning algorithms.
\end{abstract}

\begin{IEEEkeywords}
  nonholonomic mobile robot, path planning, space adaptive search
\end{IEEEkeywords}

%
\IEEEpeerreviewmaketitle

\section{Introduction}
%
%
%
%

\IEEEPARstart{C}{ompared} with holonomic mobile robots that mostly consist of very complex mechanical structures and can only be applied on flat and clean surfaces at very low speeds, the platform of nonholonomic mobile robots is much simpler, increasing the system stability and allowing the robot to move on rough terrains at higher speeds. However, due to nonholonomic constraints, the robot can only steer to limited orientations based on a given state. Therefore, for nonholonomic mobile robots, the orientation $\theta$ is required to be integrated during path planning as the third dimension of the state-space. This poses a severe challenge for path planning, since path planning under the three-dimensional $S^3  =\{x, y, \theta\}$ space requires more computation time and memory cost.

There are two major types of path planning algorithms for nonholonomic mobile robots, the randomized and deterministic algorithms. The rapidly exploring random tree (RRT) (\cite{Lavalle98rapidly_exploringrandom,LaValle99randomizedkinodynamic,Lavalle00RRT-connect}) is a widely used randomized path planning algorithm.  The RRT iteratively creates a tree object rooted on the start state $\zeta_{start}$ and explores state-space until the goal state $\zeta_{goal}$ is reached. At every step, it selects a random state $\zeta_{rand}$ and searches for the nearest state $\zeta_{i}$ of $\zeta_{rand}$ in the tree. A new state $\zeta_{i+1}$ is extended based on $\zeta_{i}$ by setting a control input that can motivate the mobile robot moving towards $\zeta_{rand}$. The RRT can quickly explore the state-space and is less likely to be trapped by the local minimum.

However, due to its randomness, the result of the RRT cannot be guaranteed to be optimal. The state $\zeta_{rand}$ that the RRT tries to reach is set randomly; therefore, the produced path can exhibit unpredictable or even very strange behavior. As a result, the produced path of the RRT cannot be directly applied, and always needs to be further optimized. This becomes an extra overhead. Given the same start and goal states, the RRT could produce very different results, thus making the results unpredictable and very difficult to control. 

In order to increase the optimality of the RRT, the RRT* (\cite{Karaman2010,Karaman2011}) is proposed. Apart from  creating a single connection to the newly inserted state $\zeta_{new}$ in each step, the RRT* also tries to connect with all states $\zeta^j\in X_{near}(\zeta_{new})$, where $X_{near}(\zeta_{new})$ is the set of states in the small neighborhood of $\zeta_{new}$. As long as the new connection helps reduce the cost of $\zeta^j$, the parent state of $\zeta^j$ will be replaced by $\zeta_{new}$. Meanwhile, all sub-states of $\zeta^j$ will also be updated based on the new cost. \cite{Karaman2011} proves the cost of the returned solution of the RRT* converges almost certainly to the optimum. However, this also comes at a price. As \cite{Jeon2011} pointed out, making a connection between two states by following nonholonomic constraints is computationally challenging, which becomes an extra overhead of the RRT*. Additionally, while the RRT is exploring the space, the number of extended states of the RRT grows, which causes the computation time of finding the nearest state to $\zeta_{rand}$ to increase step by step. For a clustered large scale space, such computation time cost becomes no longer negligible.

Dijkstra\textquoteright s algorithm (\cite{Dijkstra}) is one of the first deterministic path planning algorithms. As with the RRT, the search of Dijkstra\textquoteright s algorithm is also based on a tree object rooted on $\zeta_{start}$. However, rather than randomly explore the state-space, Dijkstra\textquoteright s  algorithm evaluates each state $\zeta$ with a path cost and iteratively extends the state of the lowest cost. The drawback of Dijkstra\textquoteright s algorithm lies in the search that universally explores the state-space, which could incur a very long computation time until the goal state $\zeta_{goal}$ is reached.

\begin{equation}\label{eq:atar}
  f(\zeta)=g(\zeta)+h(\zeta)
\end{equation}

In order to accelerate the search, the A* algorithm (\cite{astar}) introduces a heuristic term into the cost function as (\ref{eq:atar}). The cost function of the A* algorithm consists of two terms, the path cost $g(\zeta)$, and the heuristic cost $h(\zeta)$. $g(\zeta)$ is the actual path cost from $\zeta_{start}$ to $\zeta$ along the search tree, whereas $h(\zeta)$ is the estimated cost from $\zeta$ to $\zeta_{goal}$. The Euclidean distance is often used as the heuristic cost. For the states with the same path cost $g(\zeta)$, the one closer to $\zeta_{goal}$ would produce a lower $h(\zeta)$, and consequently a lower $f(\zeta)$. Since the A* algorithm always extends the node with the lowest cost $f(\zeta)$ based on (\ref{eq:atar}), the states close to $\zeta_{goal}$ will be explored first. This makes the search tree more goal oriented and exploring directly towards $\zeta_{goal}$, substantially reducing the computation cost.

\begin{figure}
  \centering
  \includegraphics[width=0.7\linewidth]{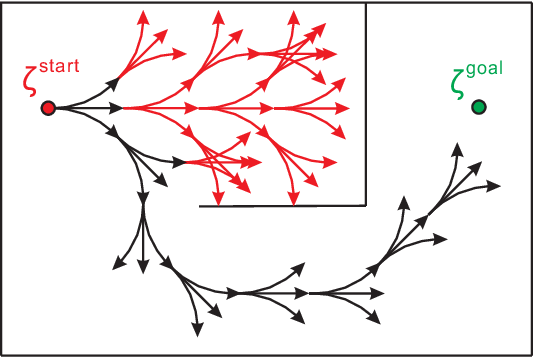}
  \caption{The process of the A* algorithm trapped by the local minimum. The red branches show the trapped steps. Only after the local minimum has been fully explored, the search can finally go towards the goal. This drawback can seriously increase the computation time and cannot be avoided by simply inflating the heuristic, such as weighted A* algorithms.}
  \label{fig:local}
\end{figure}

However the integration of the Euclidean distance-based heuristic also brings another drawback. As shown in Fig. \ref{fig:local}, if there is a concave obstacle area located between $\zeta_{start}$ and $\zeta_{goal}$, it will form a local minimum that can trap the search until the local minimum is fully explored. The red branches show the steps trapped by the local minimum during the search. Only after the local minimum is fully explored, can the search move on to explore other areas of the space shown as the black branches in Fig. \ref{fig:local}. The local minimum dramatically increases the computation cost of path planning, thus, becomes the fundamental problem of search-based path planning. 
Under a 3D state-space especially, the search could be almost indefinitely trapped. 
For real-time applications, the robot is required to find a feasible path in a very short time span. Therefore, a number of anytime algorithms were proposed, such as the Weighted A*, the Anytime A*, the Anytime Repairing A* (\cite{Bonet,Chakrabarti,Edelkamp,Korf,Likhachev,Likhachev01082009}). They are based on the fact that inflating the heuristic term may substantially speed up the search, i.e. by setting $\mu>1$ in (\ref{eq:anytime}), where $\mu$ is the heuristic coefficient. However, this is based on the sacrifice of optimality of the path. In a sense, Dijkstra and A* algorithms are special cases of anytime algorithms, where $\mu=0$ and $\mu=1$ receptively. The concept of anytime algorithms is to initially generate a suboptimal solution. If there is still time left, anytime algorithms keep improving the result until an optimal path is found or the calculating time is up. Inflating the heuristic term only helps under the environment that consists of a shallow local minimum. For environments that consist of a deep local minimum, the search could nevertheless be trapped. A state lattice planner (\cite{Pivtoraiko20051, Pivtoraiko20052, Howard2007, Pivtoraiko2009}) was proposed to improve the local planning behavior by producing precise local maneuvers, but for clustered, large-scale global path planning, the state lattice planner can still be trapped by local minima.

\begin{equation}\label{eq:anytime}
  f (\zeta) = g(\zeta) + \mu h(\zeta)
\end{equation}

Some researches try to promote the efficiency of search by providing a more accurate heuristic function $h(\zeta)$ in (\ref{eq:atar}). 
\cite{Knepper2006} proposed a heuristic look-up table to improve the local accuracy of heuristic information. However, since the table only covers a very limited neighborhood and is created ignoring obstacles, for long-range global path planning, the search with the help of a heuristic look-up table can still fall into the local minimum. Voronoi-based heuristic (\cite{qi_voronoi}) dealt with the problem of the local minimum from a different perspective. Rather than  the simple Euclidean distance, the heuristic cost is measured along the Voronoi-based roadmap that is extracted from the grid map. Voronoi-based heuristic provides a more accurate estimate to the goal. It navigates the search away from the local minimum and makes the search less likely to be trapped. However, when the search is navigated through a narrow corridor that violates nonholonomic constraints, the search could still get trapped. Such a narrow area forms another type of local minimum caused by both obstacles and nonholonomic constraints. Additionally, for partially known or even unknown environments where the robot has to create the map from scratch, it is not practical to use the Voronoi based roadmap that is extracted in advance. 

Some other works (\cite{Mirtich,Garcia,Foskey,Takahashi,Khatib,Philippsen}) try to avoid global path planning in 3D space, by applying both 2D and 3D spaces in parallel. The 2D space serves for long range global path planning ignoring nonholonomic constraints, and, therefore, the global path can be generated relatively fast. Then the global path is optimized with a short-term local path planner under the 3D space to meet nonholonomic constraints. However, the search may get trapped, when the 2D global path passes by the spot violating nonholonomic constraints. Therefore, adaptive dimensional state-spaces (\cite{zhang,Gochev}) are introduced to solve this problem. Rather than using 2D and 3D space in parallel, the state-space of variable dimensions is applied. It is assumed that most areas can be searched in 2D space. The search iteratively reconstructs the state-space by locally inserting the 3D patch at the spot where the path planning cannot be solved in 2D space. This greatly reduces the computation cost since the search can mostly be done in 2D space, and 3D path planning is only temporarily required when it is necessary. However, mixing 2D and 3D spaces together increases the complexity of the search. Furthermore, the adaptive dimensional state-space is based on a simplified environment that assumes only certain areas need benefit from the 3D planning while the vast majority of states can be searched in 2D space. Therefore, adaptive dimensionality is still not a general solution for the nonholonomic path planning problem.

In the following sections, a novel space adaptive search (SAS) approach for nonholonomic mobile robot path planning under 3D state-space is proposed. Rather than universally searching the space with fixed a step size, the steps are scaled during the search and the states are adaptively updated accordingly based on the size of the free space. This enables the search to get rid of the local minimum very fast. Under clustered large environments especially, the strength of the SAS becomes more dominant. The SAS is capable of producing a feasible path that satisfies nonholonomic constraints and only requires a very low computation cost. In order to show the strength of the SAS, we run it without heuristic later in the experiment section and compare it with the Weighted A* algorithm. It shows that, even running without heuristic, the SAS nevertheless requires much less computation time and memory cost than the Weighted A* algorithm, while still preserving the optimality of the produced path. However, the usage of the SAS is very flexible for real applications. It can be used together with heuristic or other path planning algorithms.


The problem of the motion primitive-based path planning is presented in Section 2. Section 3 proposed the SAS path planning to solve the problem, where the effective zone and scalable motion primitives are newly defined to largely speed up the search. The proposed approach is evaluated in Section 4.

\section{Motion-Primitive based Path Planning}


\begin{figure}[htbp]
  \centering
  \subfigure[]{\label{fig:state-space}\includegraphics[height=0.10\textwidth]{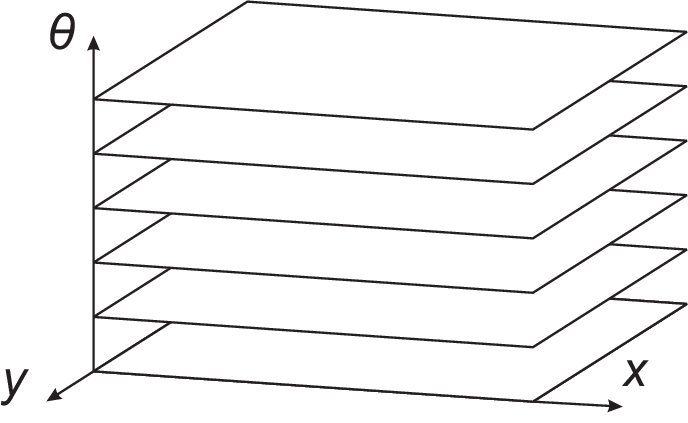}}
  \subfigure[]{\label{fig:direction_division}\includegraphics[height=0.15\textwidth]{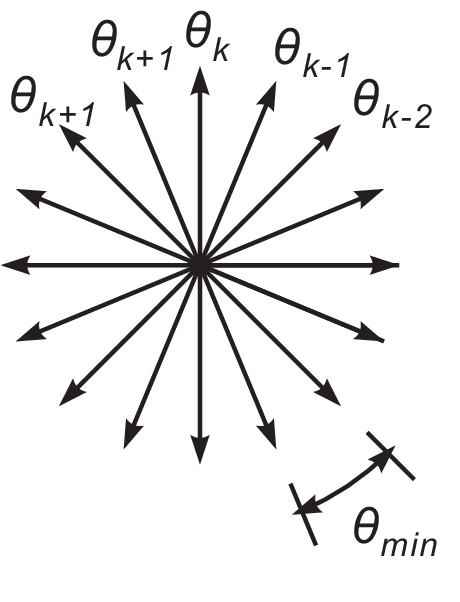}}
  \subfigure[]{\label{fig:motion_primitive}\includegraphics[height=0.15\textwidth]{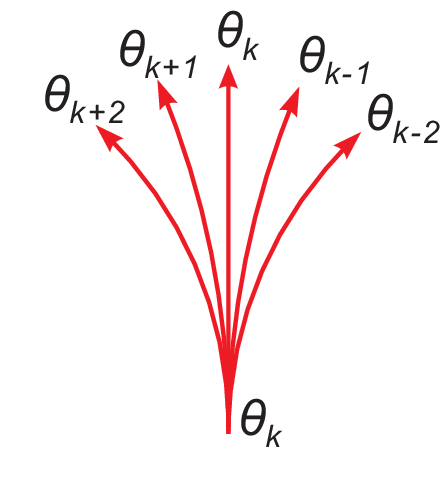}}
  \caption{(a) 3D discrete state-space $S^3  =\{x, y, \theta\}$; (b) orientation space, $\theta_{min}$ is the minimum orientation step; (c) the motion primitives defined to sequentially land in different orientations.}
\end{figure}

The motion-primitive based path planning applies a set of predefined motion primitives to iteratively build a search tree that explores the 3D state-space
$\zeta = (x, y, \theta) \in S^3$ (Fig. \ref{fig:state-space}). For simplicity, we define $u = (x, y)$, so there is $\zeta = (u, \theta)$. The orientation space $\varTheta$ is divided in K directions as in Fig. \ref{fig:direction_division} and $\theta_{min}=2\pi/K$  is the minimum orientation step, so $\varTheta$ is $\{\theta|\theta = k\theta_{min}, k \in Z\}$. The motion primitive $\tau$ defines the very basic movement that the robot can perform shortly. Fig. \ref{fig:motion_primitive} shows the concept of the motion primitives for the nonholonomic mobile robot. It is assumed that the motion primitives are similar in length and land sequentially in different orientations. With different combinations of motion primitives, more sophisticated movements can be created.

\begin{figure}[htbp]
  \centering
  \subfigure[]{\label{fig:cell_by_cell}\includegraphics[width=0.24\textwidth]{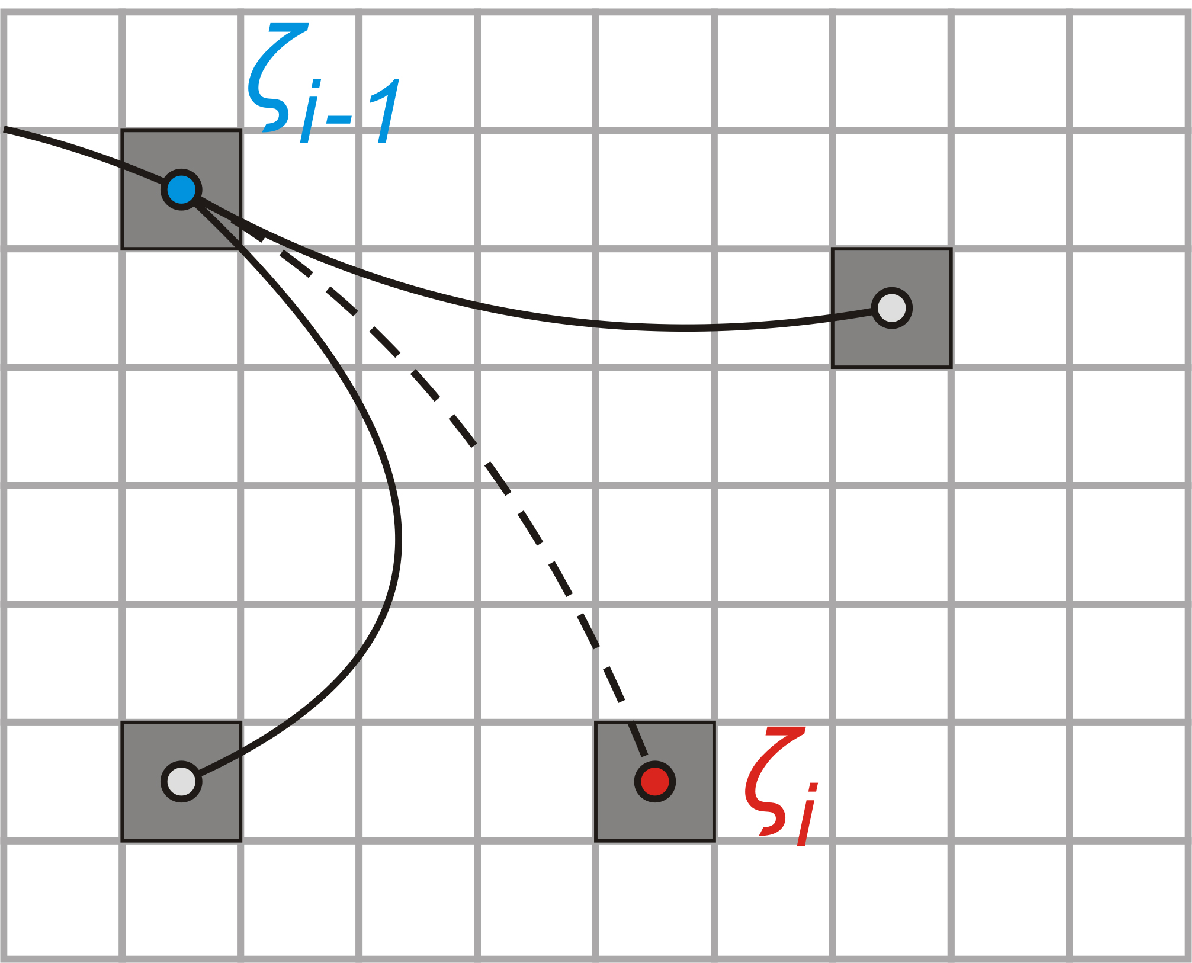}}
  \subfigure[]{\label{fig:enlaged_effective_zone}\includegraphics[width=0.24\textwidth]{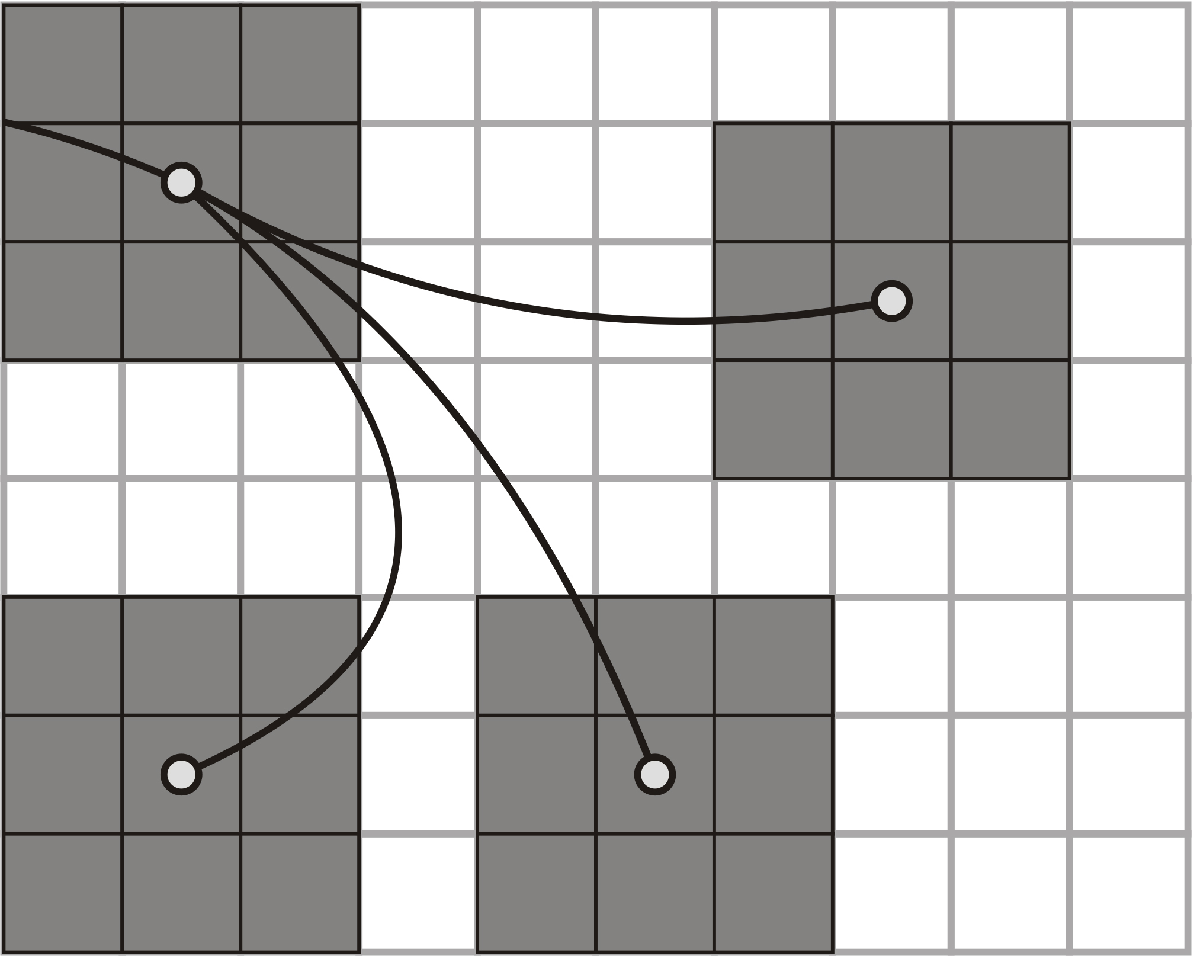}}
  \caption{(a) The classic search based path planning algorithm only update the states at the current location; (b) updating all states in the effective zone can dramatically reduce the computation time, where the effective zone is the area centered on the current state.}
\end{figure}

Fig. \ref{fig:cell_by_cell} shows the process of the motion-primitive based path planning algorithm. 
For search based path planning, the lowest path cost $g(\zeta)$ for each explored state $\zeta$ is saved under $g\_cost[\zeta]$.
$g\_cost[]$ is initiated with infinity at the beginning. Each time the search tries to create a movement from $\zeta_{i-1}$ to $\zeta_{i}$  with one of the motion primitives, the newly generated path cost $g(\zeta_i)$ will be compared with the saved path cost $g\_cost[\zeta_i]$. If there is $g(\zeta_i) < g\_cost[\zeta_i]$, then $g\_cost[\zeta_i]$ will be updated with $g(\zeta_i)$; otherwise the movement from $\zeta_{i-1}$ to $\zeta_{i}$  will be ignored. $g\_cost[]$ records the explored space, which guarantees the same state $\zeta_i$ will not be explored again, unless the value of $g\_cost[\zeta_i]$ can be reduced. This guarantees that the search iteratively explores the whole space by ignoring the states that have been explored before.

However, only to update $g\_cost[]$ of the states at the current location (shown as the dark cell in Fig. \ref{fig:cell_by_cell}) is far too slow for large-scale environment. In order to accelerate the search, as shown in Fig. \ref{fig:enlaged_effective_zone}, every time the search tries to update $g\_cost[]$ for a certain $\zeta_i$, all states $\zeta_n$ in the small neighborhood of $\zeta_i$ will all be updated with $g(\zeta_i)$ as long as $g(\zeta_i) < g\_cost[\zeta_n]$. The small neighborhood of $\zeta_i$ is defined as the effective zone $\varOmega_i$ of $\zeta_i$. Clearly, a large effective zone can speed up the search, but the narrow corridors could be thus ignored, which reduces the precision of the search; on the other hand, small effective zone can increase the precision, but will slow down the search.

\section{Space Adaptive Search}	
In order to achieve both high precision and efficiency, the space adaptive search (SAS) is proposed in the following sections, which dynamically changes the size of the effective zone during the search.

\subsection{Effective Zone}\label{sec:effective}
\begin{figure*}[htbp]
  \centering
  \subfigure[]{\label{fig:clear_space}\includegraphics[width=0.4\textwidth]{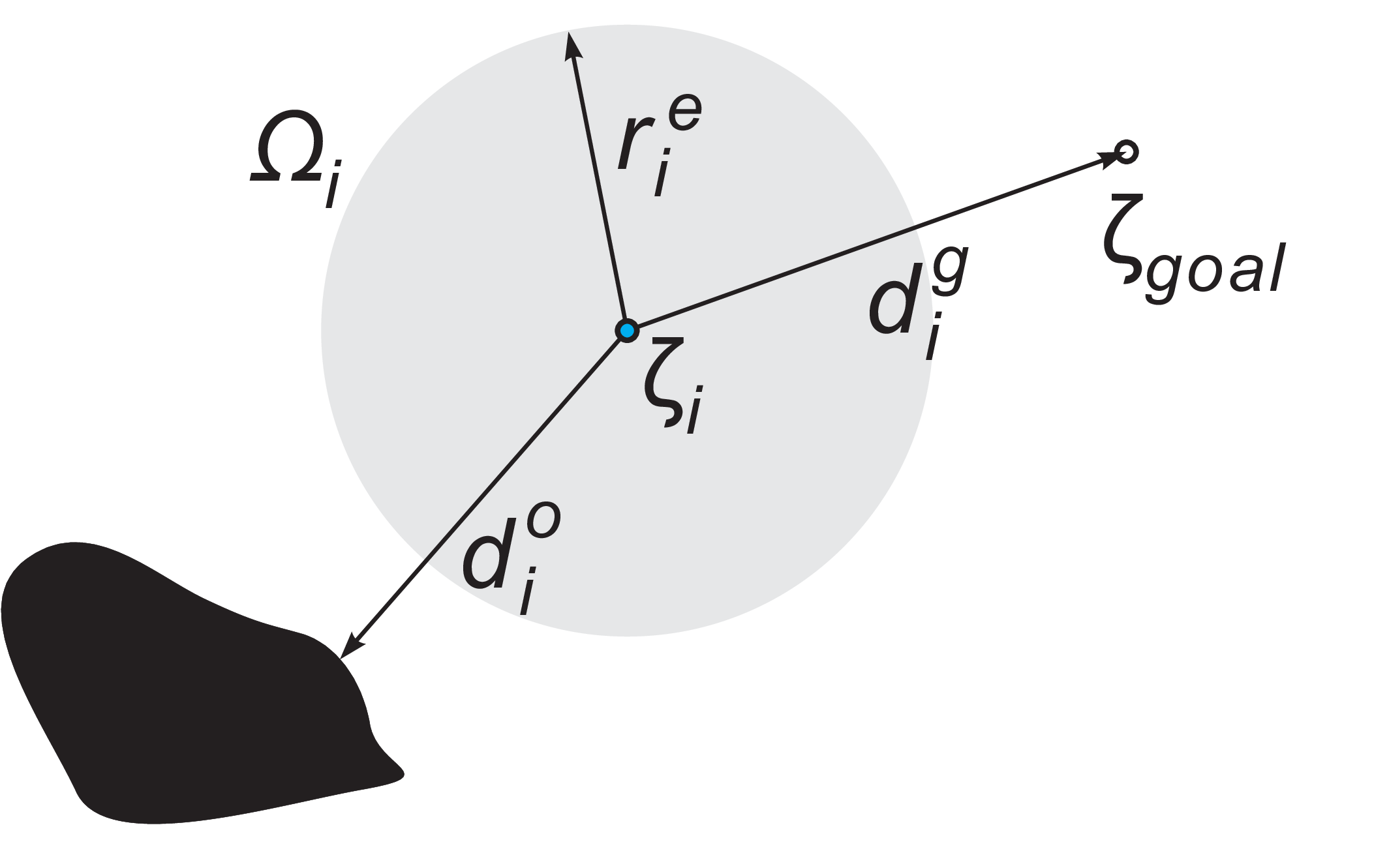}}
  \subfigure[]{\label{fig:effective_zone}\includegraphics[width=0.4\textwidth]{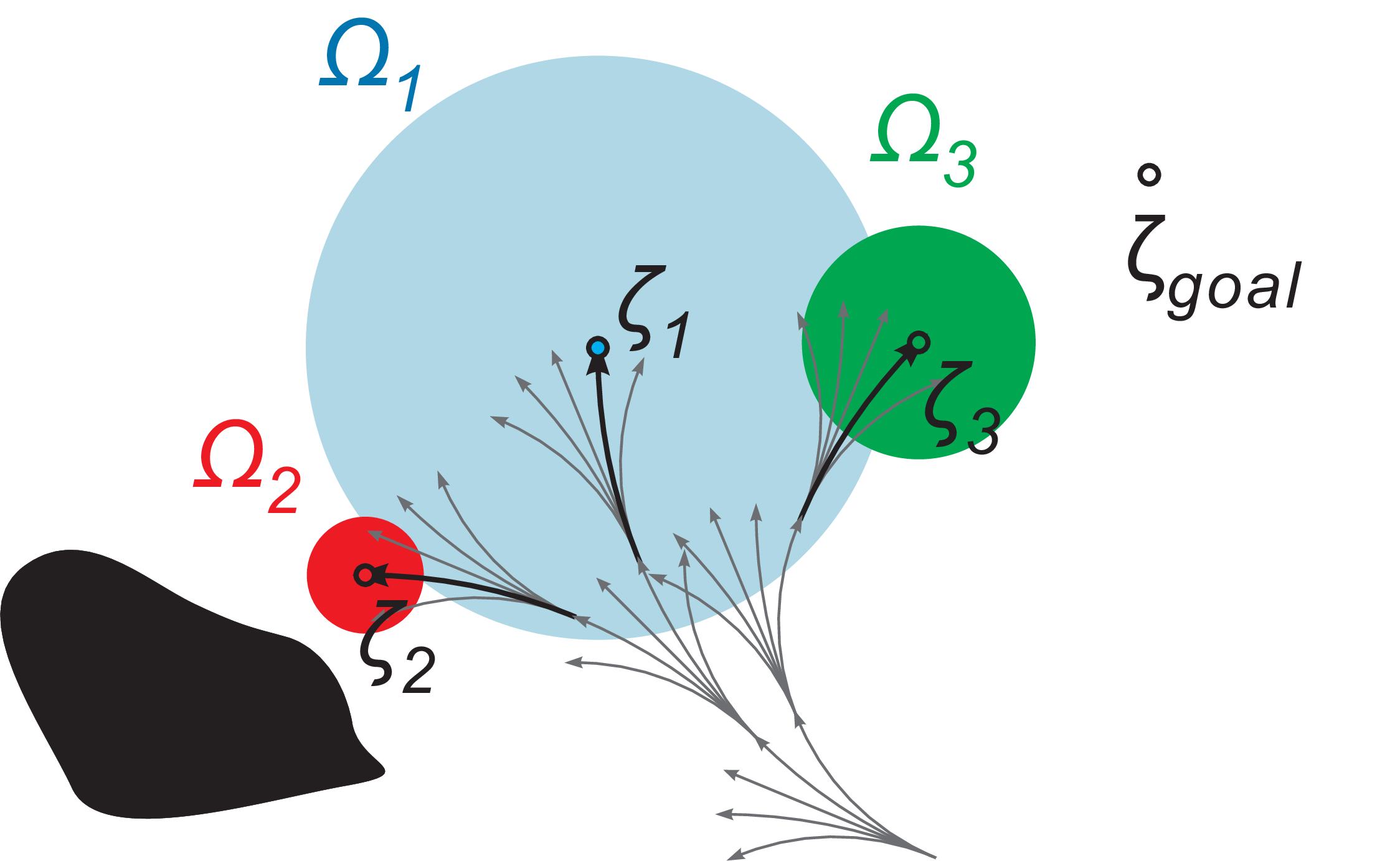}}
  \caption{(a) The radius $r_i^e$ of the effective zone $\varOmega_i$ of $\zeta_i$ depends on the distance to its closest obstacle $d^o_i$ and its distance to the goal $d^g_i$; (b) the effective zone shrinks while the search is approaching the obstacle or the goal.}
\end{figure*}

As shown in Fig. \ref{fig:clear_space}, the effective zone $\varOmega_i$ of $\zeta_i$ is a circular area centered on $\zeta_i$, which represents an obstacle-free space around $\zeta_i$. As defined by (\ref{eq:effective_zone}), where $r^e_i$ is the radius of $\varOmega_i$, $u_i$ and $\theta_i$ are the position and orientation of $\zeta_i$ respectively, and $dist(u_n, u_i)$ is the Euclidean distance between $u_n$ and $u_i$. Note that $\varOmega_i$ is defined as a set of states in the same orientation with  $\zeta_i=(u_i,\theta_i)$.

\begin{equation}\label{eq:effective_zone}
  \varOmega_i=\{\zeta_n=(u_n,\theta_n) |dist(u_n , u_i) < r^e_i\text{\ \ } \wedge\text{\ \ } \theta_n = \theta_i, \zeta_n \in S^3\}
\end{equation} 

As in Fig. \ref{fig:clear_space}, $d^o_i$  is the Euclidean distance between $\zeta_i$  and its closest obstacle, $d^g_i$ is the Euclidean distance between $\zeta_i$ and the goal. The concept of the SAS based path planning is simple, the areas close to the obstacle or the goal need to be more precisely searched. Those areas are the critical places that pose more potential risks of violating the nonholonomic constraints, thus, determine whether or not the path can be successfully found. Therefore, the size of $\varOmega_i$ is reduced, when $\zeta_i$ lands close to the obstacle or the goal. To achieve this, $r^e_i$ is set as the minimum value between $\kappa^o d^o_i$  and $\kappa^g d^g_i$  as (\ref{eq:effective_radius}). $\kappa^o$ and $\kappa^g$ are the coefficients used to tune the effect of $d^o_i$ and $d^g_i$. It is observable that both $d^o_i$ and $d^g_i$ are not allowed to be greater than one, this guarantees that $\varOmega_i$ will not cover obstacles or the goal.

\begin{equation}\label{eq:effective_radius}
  r_i^e=min(\kappa^o d^o_i , \kappa^g d^g_i)\text{\ }(1\geq\kappa^g>0, 1\geq\kappa^o> 0)
\end{equation}

As in Fig. \ref{fig:effective_zone}, $\varOmega_1$, $\varOmega_2$  and $\varOmega_3$  are the effective zones of $\zeta_1$, $\zeta_2$  and $\zeta_3$ respectively.  It is observable that $\varOmega_1$  is larger than both $\varOmega_2$  and $\varOmega_3$.  This is because $\zeta_1$ is far from both the obstacles and the goal, and there is more free space around $\zeta_1$. The precision of the search around $\zeta_1$ becomes less important, so the effective zone of $\zeta_1$ is set relatively large. However, $\zeta_2$ is close to the obstacle and $\zeta_3$ is close to the goal, and the spaces around $\zeta_2$ and $\zeta_3$ need to be more precisely searched, so $\varOmega_2$ and $\varOmega_3$ are set very small. Note that since $\zeta_1$, $\zeta_2$ and $\zeta_3$ land in different orientations, $\varOmega_1$, $\varOmega_2$ and $\varOmega_3$ are three sets of states that do not overlap each other.

It is also observable in Fig. \ref{fig:clear_space} that the effective zone only depends on $d^o$ and $d^g$, and has nothing to do with the resolution of the map. The SAS will update all the states in the effective zone in one step, as long as $g(\zeta_i) < g\_cost[\zeta_n]$. Therefore, the search becomes relatively independent from the resolution of the map. This is a great feature of the SAS that the environment can be represented with a relatively high resolution map, without dramatically increasing the computation cost.

\begin{figure}[htbp]
  \centering
  \subfigure[]{\label{fig:grid_map}\includegraphics[width=0.2\textwidth]{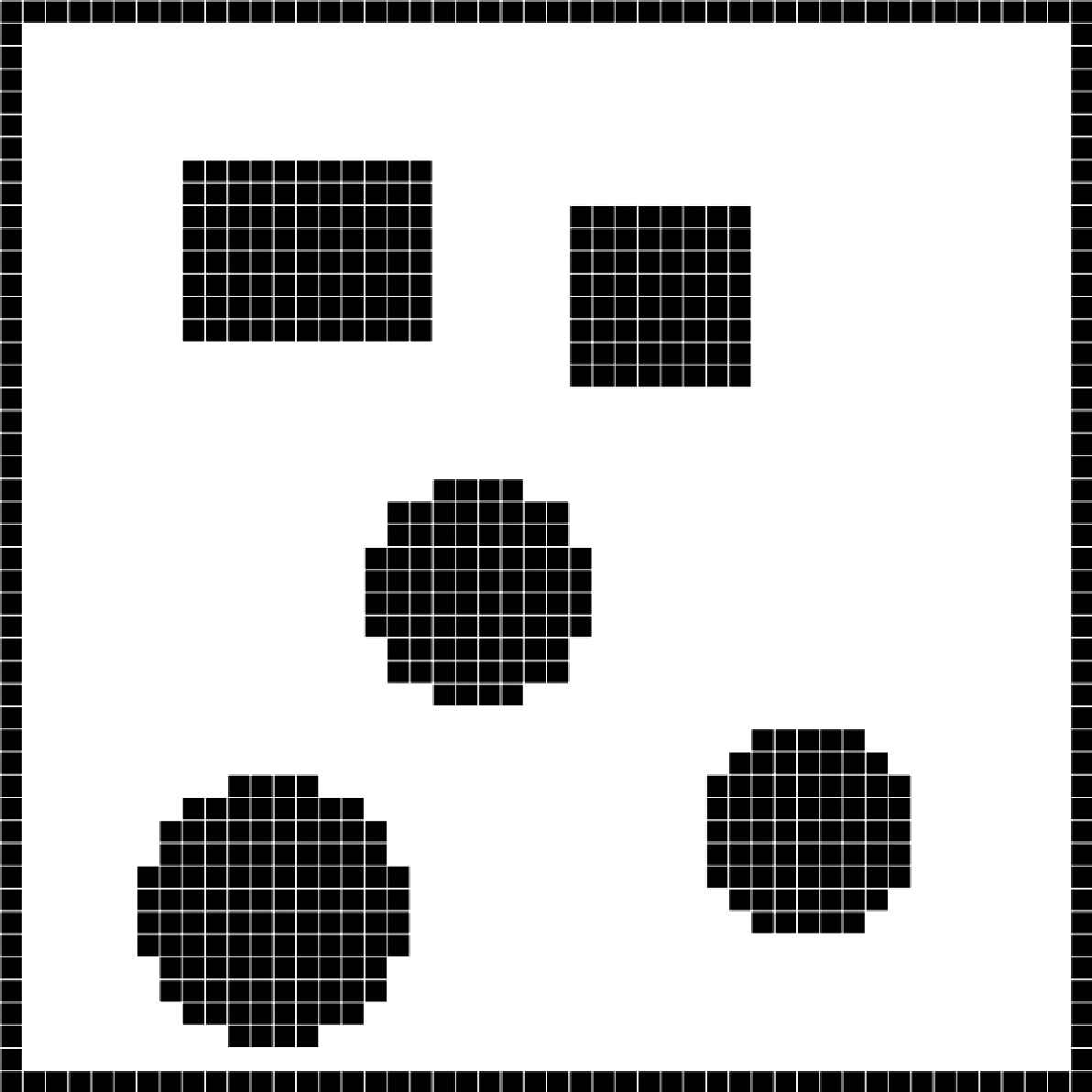}}
  \subfigure[]{\label{fig:dist_map}\includegraphics[width=0.2\textwidth]{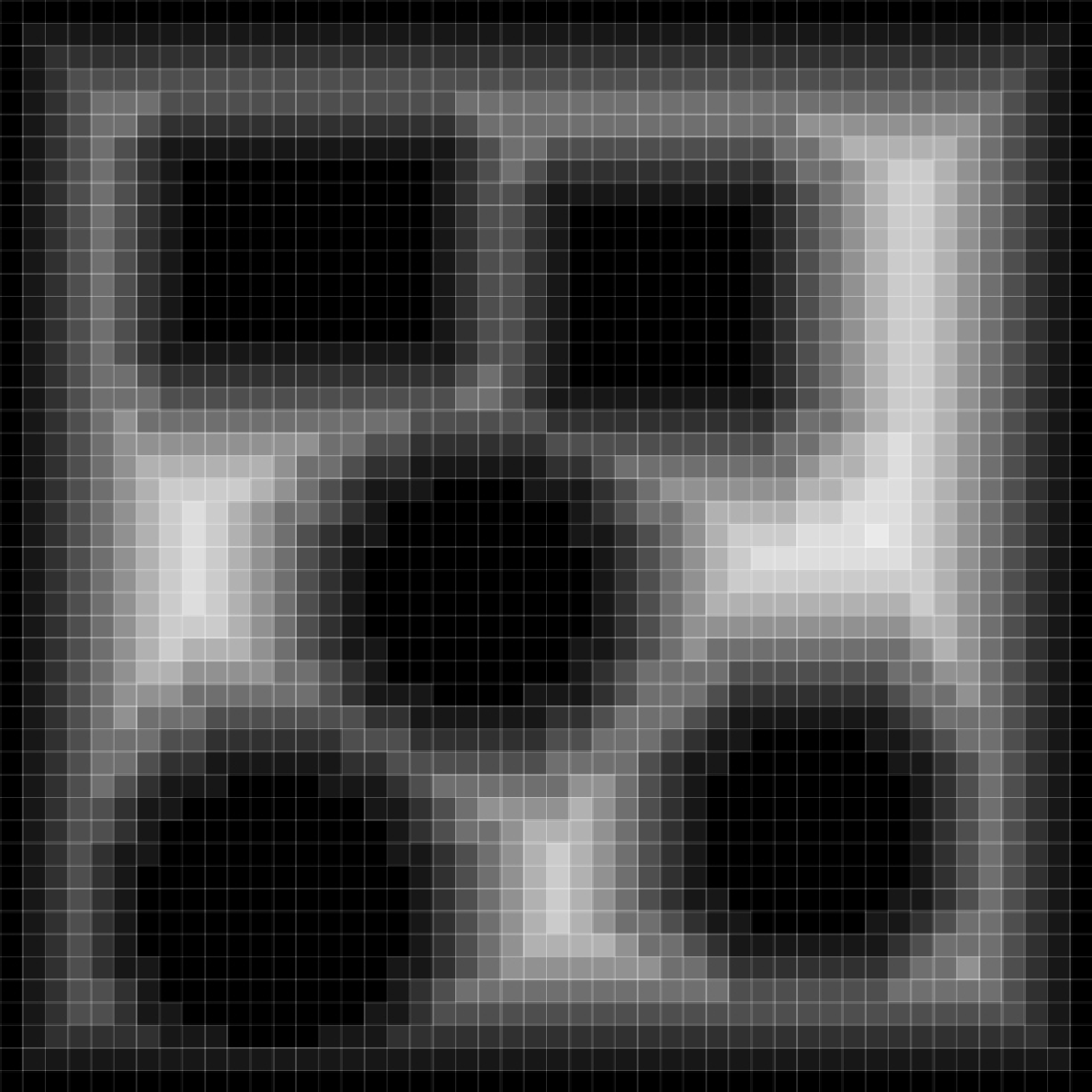}}
  \caption{(a) Grid map: the black cells represent the locations of obstacles; (b) the clearance map is transformed from the grid map with the distance transformation. Each cell is attached with a value representing the distance to its closest obstacle. This can be used as a measurement of the free space around the current location.}
\end{figure}

$d^g_i$ is simply the distance to the goal, whereas the value of $d^o_i$ can be obtained from the clearance map.  Fig. \ref{fig:grid_map} shows a grid map, where the black cells represent the location of the obstacles. Fig. \ref{fig:dist_map} is the clearance map transformed from the grid map with distance transformation (\cite{Shih, Cuisenaire}). In the clearance map, each cell is attached with a value that represents the distance to its closest obstacle.

In short, the proposed SAS path planning will roughly explore the area that possesses more free space and precisely explore the area relatively close to the obstacle and the goal. Observably, reducing both $d^g$ and $d^o$ in (\ref{eq:effective_radius}) to zero can make the effective zone extremely small, so the search will only update the state on the current location. In this case, the SAS becomes Dijkstra's algorithm.

\subsection{Ideal Motion Primitives}	
\begin{figure*}[htbp]
  \centering
  \subfigure[]{\label{fig:small_steps}\includegraphics[width=0.25\textwidth]{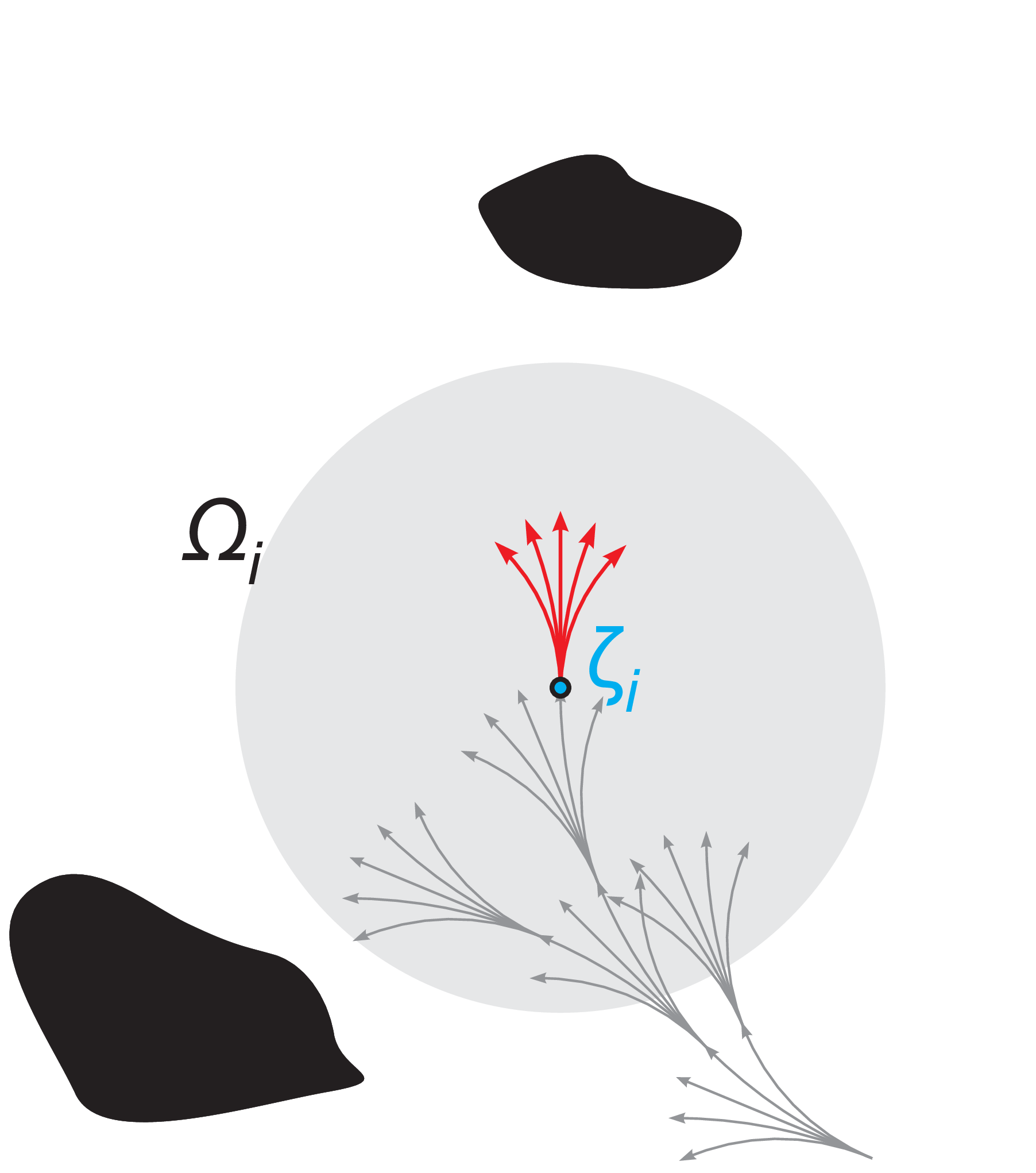}}\hspace{0.1\textwidth}
  \subfigure[]{\label{fig:over_large_steps}\includegraphics[width=0.25\textwidth]{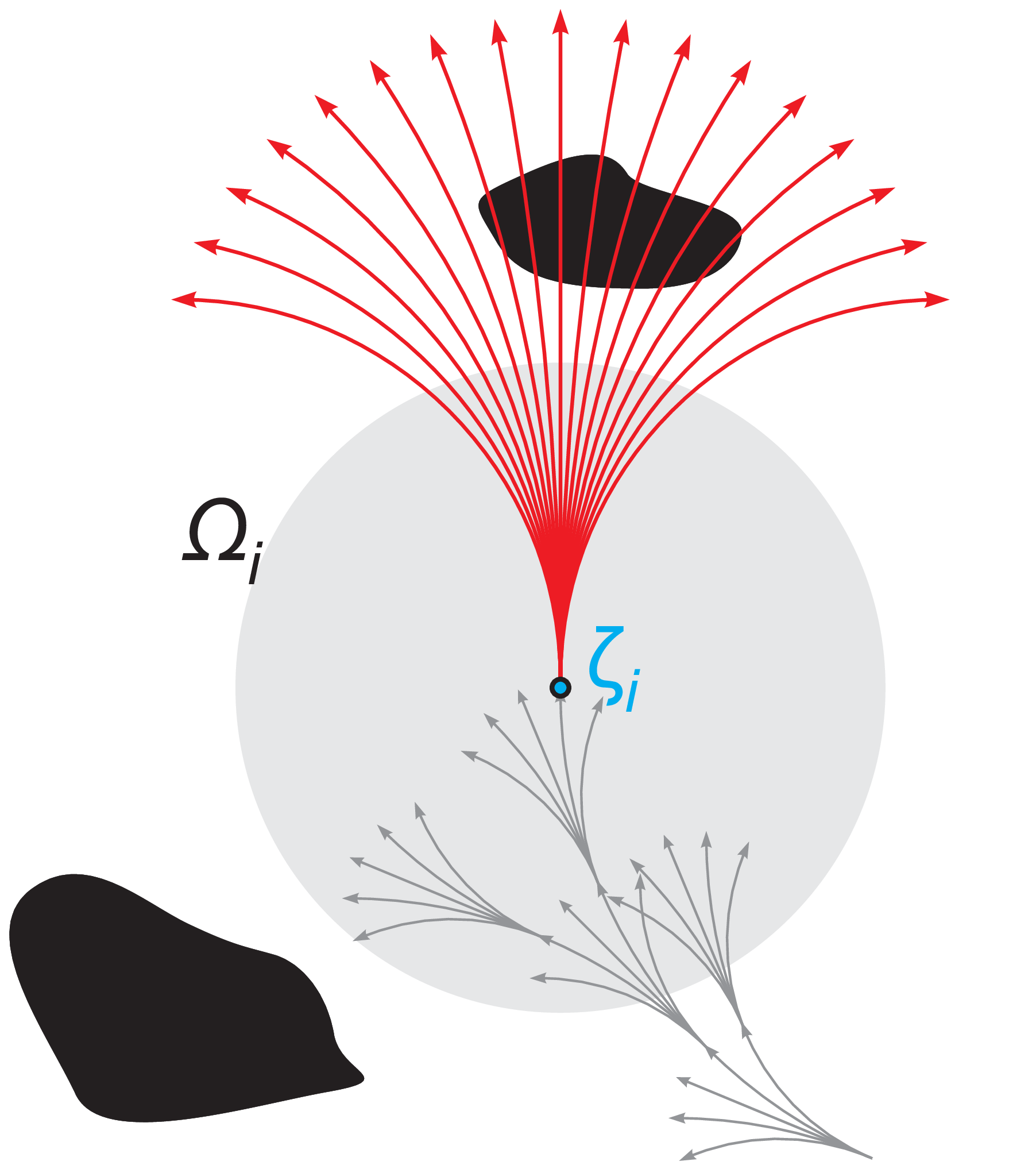}}\hspace{0.1\textwidth}
  \subfigure[]{\label{fig:ideal_steps}\includegraphics[width=0.25\textwidth]{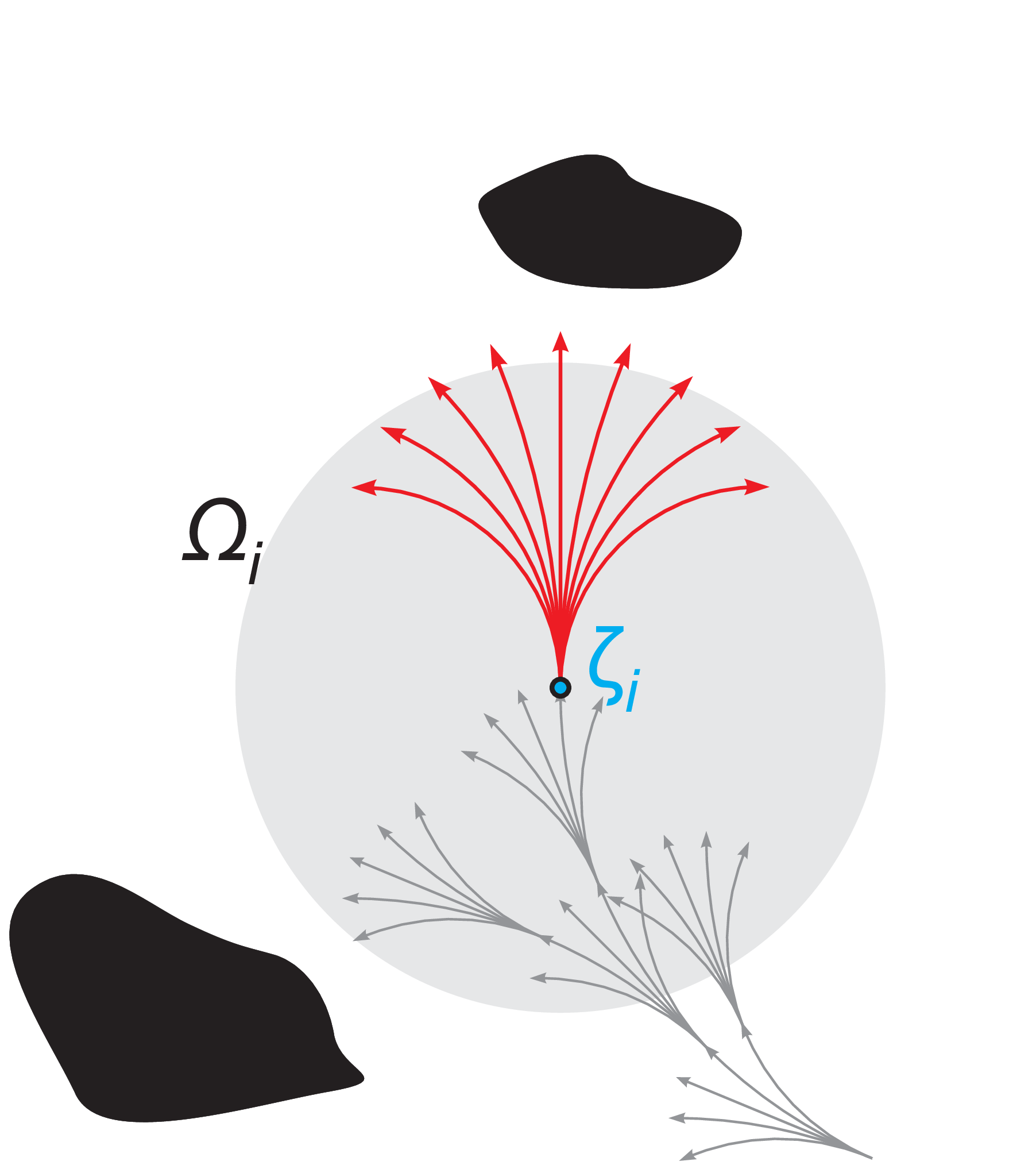}}
  \caption{(a) The step is too small, which may cause the following nodes to be trapped in the effective zone of the current node; (b) the step is too large, which increases the risk of the following steps being blocked by obstacles; (c) the steps are expected to go slightly beyond the effective zone.}
  \label{fig:motion_primitive_steps}
\end{figure*}

Fig. \ref{fig:motion_primitive_steps} shows several types of motion primitives in different sizes. $\zeta_i$ is one of the states in the exploring tree and the red sub-tree shows the successive steps that expand $\zeta_i$. The gray circular area is the effective zone $\varOmega_i$ of $\zeta_i$.

In Fig. \ref{fig:small_steps}, compared with $\varOmega_i$, the successive steps of $\zeta_i$ are relatively too small, which leads them to be covered in $\varOmega_i$. Since all states in $\varOmega_i$ are updated with $g(\zeta_i)$, it could lead successive steps of $\zeta_i$ to be ignored. Furthermore, if there is a large free space around $\zeta_i$, short exploring steps will make the search very inefficient.
In Fig. \ref{fig:over_large_steps}, the successive steps of $\zeta_i$ are too large, which go too far beyond $\varOmega_i$. This increases the risk of the successive steps being blocked by the obstacles, and consequently produces more failed steps that make the search less efficient.

As in Fig. \ref{fig:ideal_steps}, the ideal motion primitives are expected to go slightly beyond $\varOmega_i$, which make the steps go far enough without increasing the risk of being blocked by the obstacles. It seems that some of the motion primitives are still enclosed by $\varOmega_i$. However, those motion primitives are ended in a different orientation from $\zeta_i$, and will, therefore, not be covered by $\varOmega_i$. 

\subsection{Scalable Motion Primitives} 
\begin{figure}[htbp]
  \centering
  \subfigure[]{\label{fig:steerings}\includegraphics[height=0.3\textwidth]{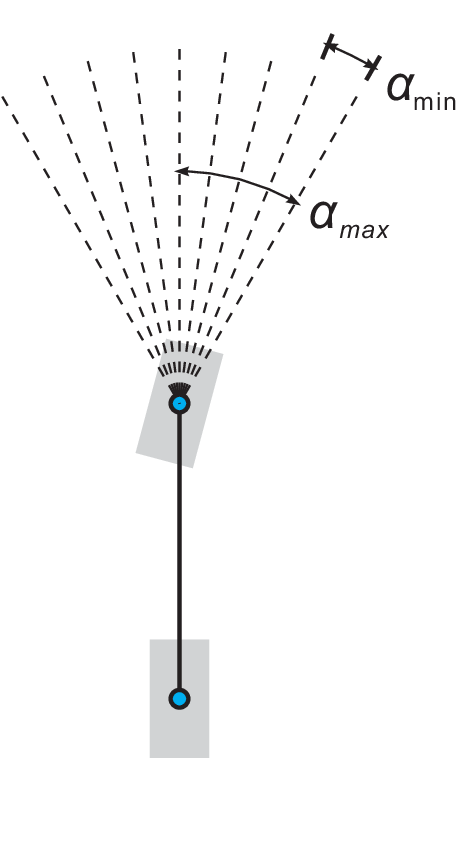}}
  \subfigure[]{\label{fig:geometry}\includegraphics[height=0.3\textwidth]{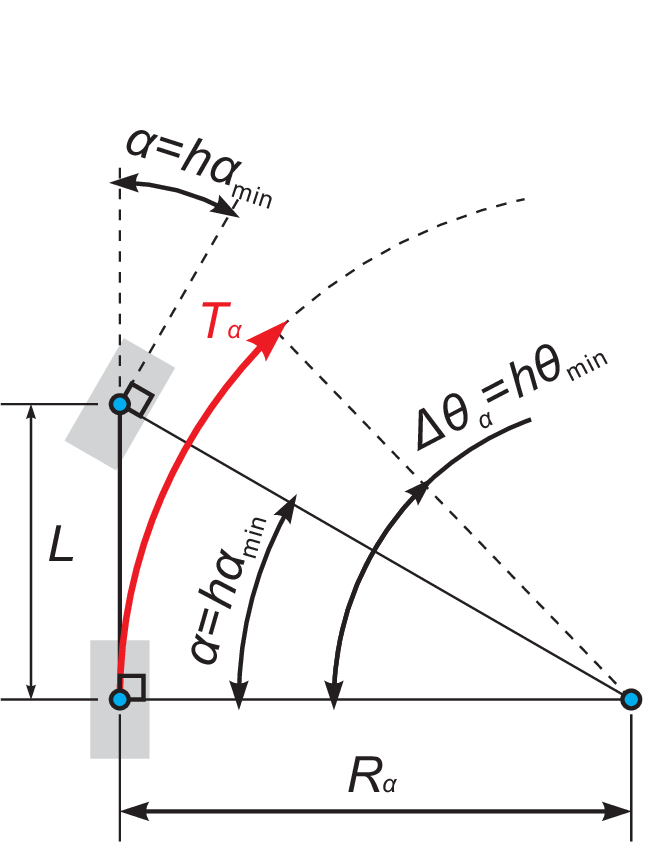}}
  \caption{(a) The steering space is discretized in $H$ sections, the minimum steering step is $\alpha_{min}$; (b) the orientation of motion primitive $T_{\alpha}$ turns $\varDelta \theta_{\alpha} = h\theta_{min}$  by applying the steering angle $\alpha = h\alpha_{min}$ on the front wheel.} 
  \label{fig:bicycle_model}
\end{figure}

Fig. \ref{fig:bicycle_model} shows a simplified bicycle model of a vehicle (\cite{ground_vehicle_dynamics}). The front and rear wheels are represented as the virtual wheels in the center respectively (shown as the gray boxes in Fig. \ref{fig:bicycle_model}). The wheelbase is set as $L$. The steering space is divided in $H$ sections, and the minimum steering angle is $\alpha_{min}$.

As in Fig. \ref{fig:geometry}, it is defined that for the primitive trajectory $T_{\alpha}$ with the steering angle $\alpha = h\alpha_{min}$ ($h\in \mathbb{Z}$), $T_{\alpha}$ terminates when the orientation of $T_{\alpha}$ turns $\varDelta \theta_{\alpha} = h\theta_{min}$. Based on Fig. \ref{fig:geometry}, $\varDelta s_{\alpha}$ represents the length of $T_{\alpha}$, and it can be calculated with (\ref{eq:alpha_length}).

\begin{equation}\label{eq:alpha_length}
  \begin{array}{r}
    \varDelta s_{\alpha}=R_{\alpha}\Delta\theta_{\alpha}
    =\dfrac{L}{tan(\alpha)}h\theta_{min}
    =\dfrac{L}{tan(h\alpha_{min})}h\theta_{min}\\
    \\
    \text{\hspace{0.1\textwidth}}
    (\alpha=h\alpha_{min}, h\neq 0)
  \end{array}
\end{equation}

However, when the steering angle $\alpha = 0 (h = 0)$, there are $R_{\alpha} = \infty$ and $\varDelta\theta_{\alpha} = 0$, which make the length $\varDelta s_0$ of the trajectory $T_{\alpha}(\alpha = 0)$ not directly calculable with (\ref{eq:alpha_length}) by setting $h = 0$. Therefore, we substitute $h$ in (\ref{eq:alpha_length}) with an indefinitely small value $\epsilon \in \mathbb{R}$, then the calculation of $\varDelta s_0$ could be done by taking the limit of $\varDelta\theta_{\alpha}$ with $\epsilon \rightarrow 0$ as (\ref{eq:zero_steer_limit}). When $\epsilon \rightarrow 0$, then $tan(\epsilon\alpha_{min}) \rightarrow \epsilon\alpha_{min}$, so $tan(\epsilon\alpha_{min})$ in (\ref{eq:zero_steer_limit}) can be substituted with $\epsilon\alpha_{min}$, which yields (\ref{eq:zero_steer_limit_2}).

\begin{equation}\label{eq:zero_steer_limit}
  \varDelta s_{0}
  =\lim_{\alpha \to 0} \varDelta s_{\alpha}
  =\lim_{\epsilon \to 0}\dfrac{L}{tan(\epsilon\alpha_{min})}\epsilon\theta_{min}
\end{equation}

\begin{equation}\label{eq:zero_steer_limit_2}
  \varDelta s_{0}
  =\lim_{\epsilon \to 0}\dfrac{L}{\epsilon\alpha_{min}}\epsilon\theta_{min}
  =\dfrac{L\theta_{min}}{\alpha_{min}}
\end{equation}

\begin{figure}[htbp]
  \centering
  \subfigure[]{\label{fig:extendable_primitive}\includegraphics[height=0.17\textwidth]{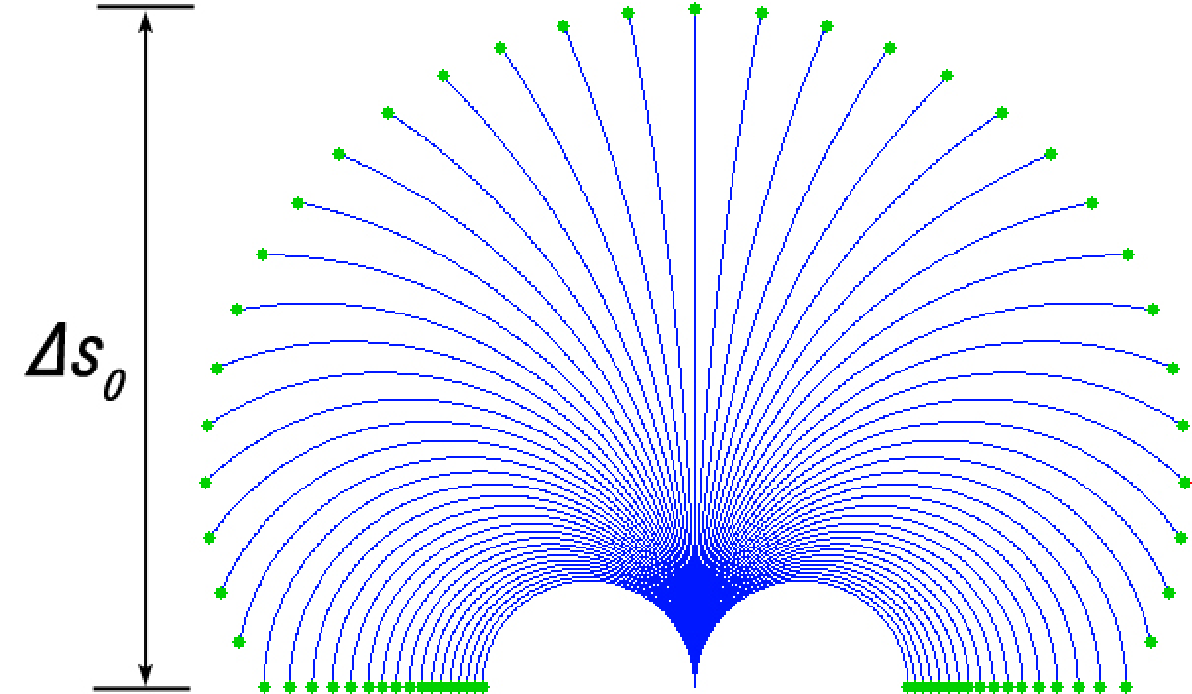}}
  \subfigure[]{\label{fig:scaled_primitive}\includegraphics[height=0.17\textwidth]{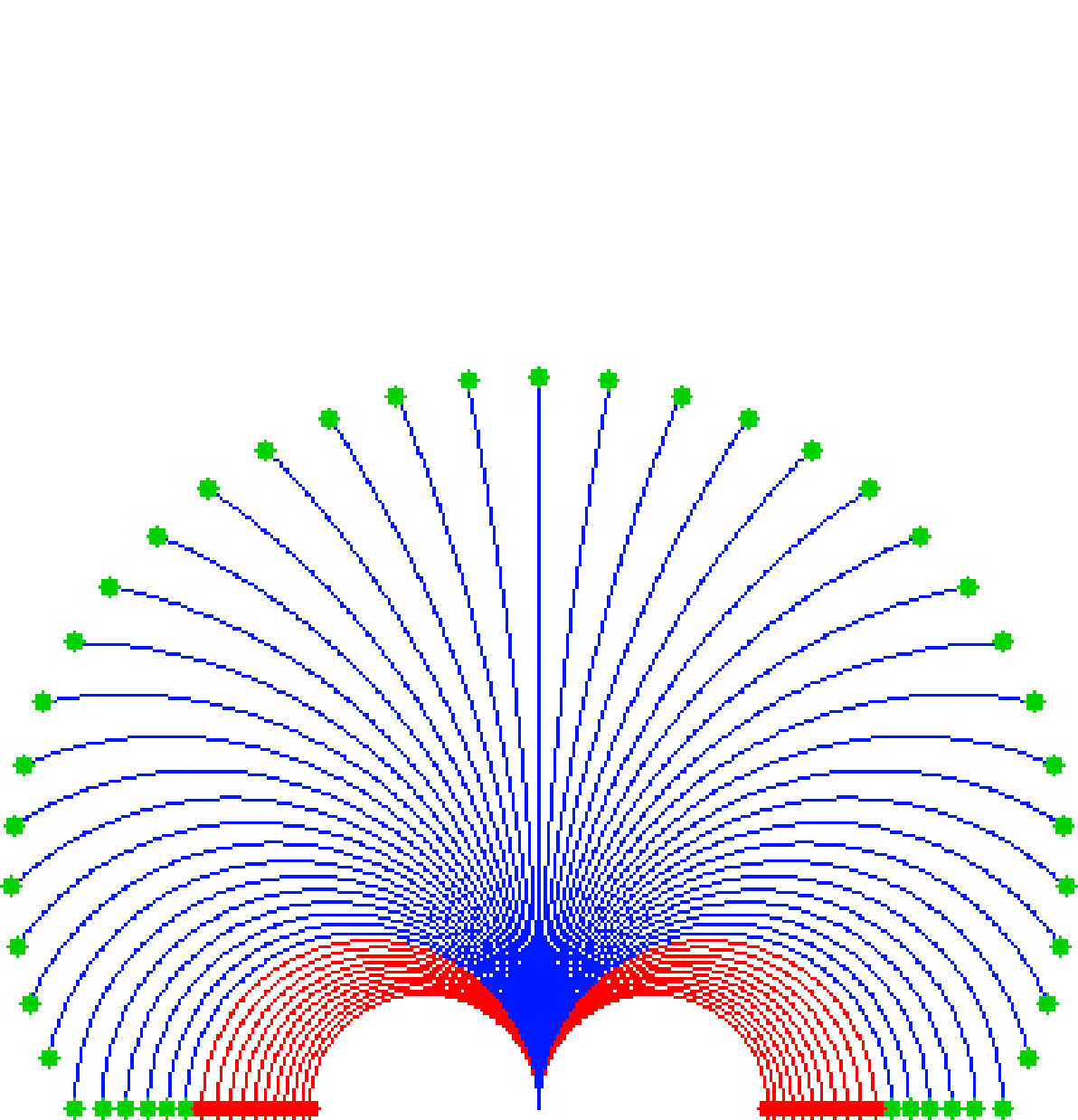}}
  \caption{(a) Motion primitives, $K = 32$, $H = 64$; (b) scaled motion primitives with $\eta_i = 2/3$, the red trajectories show the motion primitives that violate nonholonomic constraints after scaling down.
  }
\end{figure}

Fig. \ref{fig:extendable_primitive} shows the motion primitives, by setting the division of the orientation space $K = 32$ and the steering space $H = 64$. However, the motion primitives are not allowed to turn over $\pi$, so when $h\theta_{min} > \pi$ , there is $\varDelta \theta_{\alpha} = \pi$.  
So, there are some motion primitives that stop at the bottom line, where $\varDelta\theta = \pi$. $\varDelta s_0$ is the length of the motion primitive $T_{\alpha}$ of 0 steering angle, which is also the motion primitive with maximum length.

As mentioned, the motion primitives should slightly go beyond the effective zone $\varOmega_i$ of $\zeta_i$ as in Fig. \ref{fig:ideal_steps}. This can be achieved by simply scaling down the motion primitives with a scaling factor $\eta_i\leq 1$  as (\ref{eq:scalar}), where $r^e_i$ is the radius of $\varOmega_i$. $\lambda$ is the minimum step that the mobile robot can physically perform. We can see that $\lambda$ always makes sure that the scaled motion primitives go slightly beyond $\varOmega_i$.

\begin{equation}\label{eq:scalar}
  \eta_i =\dfrac{r^e_i + \lambda}{\varDelta s_0}
\end{equation}

\begin{equation}\label{eq:delta_s_constraints}
  r^e_i\leq \varDelta s_0-\lambda
\end{equation}

Based on (\ref{eq:scalar}), to ensure the motion primitives always go beyond the effective zone ($\eta_i\leq 1$), the radius $r^e_i$ of the effective zone must meet (\ref{eq:delta_s_constraints}). Therefore (\ref{eq:delta_s_constraints}) becomes another constraint of $r^e_i$. As a result, the definition of $r^e_i$ defined by(\ref{eq:effective_radius}) is now transformed into (\ref{eq:effective_radius_final}). In order to reduce the constraint of (\ref{eq:delta_s_constraints}), $\varDelta s_0$ should be increased as much as possible. According to (\ref{eq:zero_steer_limit_2}), this can be achieved by either reducing $\alpha_{min}$, or increasing $\theta_{min}$. However, increasing $\theta_{min}$ reduces the resolution of the orientation dimension $\theta$, which will reduce the flexibility of the search. Therefore reducing $\alpha_{min}$ is the better choice.

\begin{equation}\label{eq:effective_radius_final}
  \begin{array}{r}
    r_i^e=min(\kappa^o d^o_i ,\ \kappa^g d^g_i,\ \varDelta s_0-\lambda)\\
    \\
    (1\geq\kappa^g>0, 1\geq\kappa^o> 0)
  \end{array}
\end{equation}

Scaling down the motion primitives can produce new types of motion primitives with various sizes and curvatures. This produces a path in a more general form, and no longer limited by the predefined motion primitives. Fig. \ref{fig:scaled_primitive} is the scaled motion primitives in Fig. \ref{fig:extendable_primitive} with a scaling factor $\eta_i=2/3$. It is observable that scaling-down could increase the curvature of motion primitives. This makes the curvatures of some scaled motion primitives become too large, thus, violate nonholonomic constraints, as shown by red trajectories in Fig. \ref{fig:extendable_primitive}. The SAS will only apply the scaled motion primitives that follow the nonholonomic constraints during the path planning process, the ones (the red trajectories in Fig. \ref{fig:extendable_primitive}) violating nonholonomic constraints will be ignored. For very clustered, narrow environments, additional motion primitive sets with a smaller scaling factor can also be used together at every step.

\section{Experiments}
To demonstrate the usefulness of the SAS, it is compared with the Weighted A* algorithm. We run SAS without heuristic, whereas the Weighted A* algorithm runs with Euclidean distance-based heuristic. The Weighted A* applies a set of motion primitives of constant lengths and updates the states at the current location in each step. Both the SAS and the Weighted A* are evaluated by utilizing the same path cost $g()$. The path cost $g()$ is defined based on the length of the path. 
The definitions of cost functions are alternative for the SAS, but this is out of the scope of this paper. The test is programmed with c++ and runs under a 64 bit windows 7 operating system with Intel i7-4770K CPU 3.50 GHz.

Three different grid maps, A [Fig. \ref{fig:cpa_local_path}], B [Fig. \ref{fig:ccs_close_to_goal_path}] and C [Fig. \ref{fig:ccs_normal_path}] are tested to show the the SAS's strength from different perspectives. The red and the green arrows represent the start and the goal respectively. The first experiment in Section \ref{sec:local_minimum} shows the general idea of how the SAS effectively gets rid of the local minimum. Two coefficients $\kappa^g$ and $\kappa^o$ of (\ref{eq:effective_radius_final}) that affect the SAS are tested in Sections \ref{sec:goal} and \ref{sec:general}. Section \ref{sec:goal} shows that the SAS converges to the goal very fast, and how it is affected by $\kappa^g$. Section \ref{sec:general} shows the test of the SAS under a more general environment, and how $\kappa^o$ can affect the path planning process.

Table \ref{tab:maps} shows the size of grid maps and the computation cost of distance transformation introduced by \cite{Cuisenaire} for each grid map.  
As mentioned in section \ref{sec:effective}, distance transformation is required by SAS to measure $d^o_i$, i.e. the distance from the current state to its closest obstacle. As Table \ref{tab:maps} shows, the computation time of distance transformation only require a few milliseconds.

\begin{table}[t]
  \caption{Computation time of the distance transformation applied on the map of different size}\label{tab:maps}
  \begin{tabular*}{0.48\textwidth}{@{\extracolsep\fill}llll}
    \hline
    Map	& Map Size			& Time Cost of Distance Transformation\\
    \hline 
    A (Fig. \ref{fig:cpa_local_path})	&$350\times 200$ 	& 0.96 ms 		\\

    B (Fig. \ref{fig:ccs_close_to_goal_path})	&$500\times 300$ 	& 0.91 ms 		\\

    C (Fig. \ref{fig:ccs_normal_path})	&$1000\times 500$ 	& 9.73 ms 		\\
    \hline
  \end{tabular*}
\end{table}

\subsection{Deep Local Minimum}\label{sec:local_minimum}
\begin{figure*}[htbp]
  \centering
  \subfigure[Produced path]{\label{fig:cpa_local_path}\includegraphics[width=0.24\textwidth]{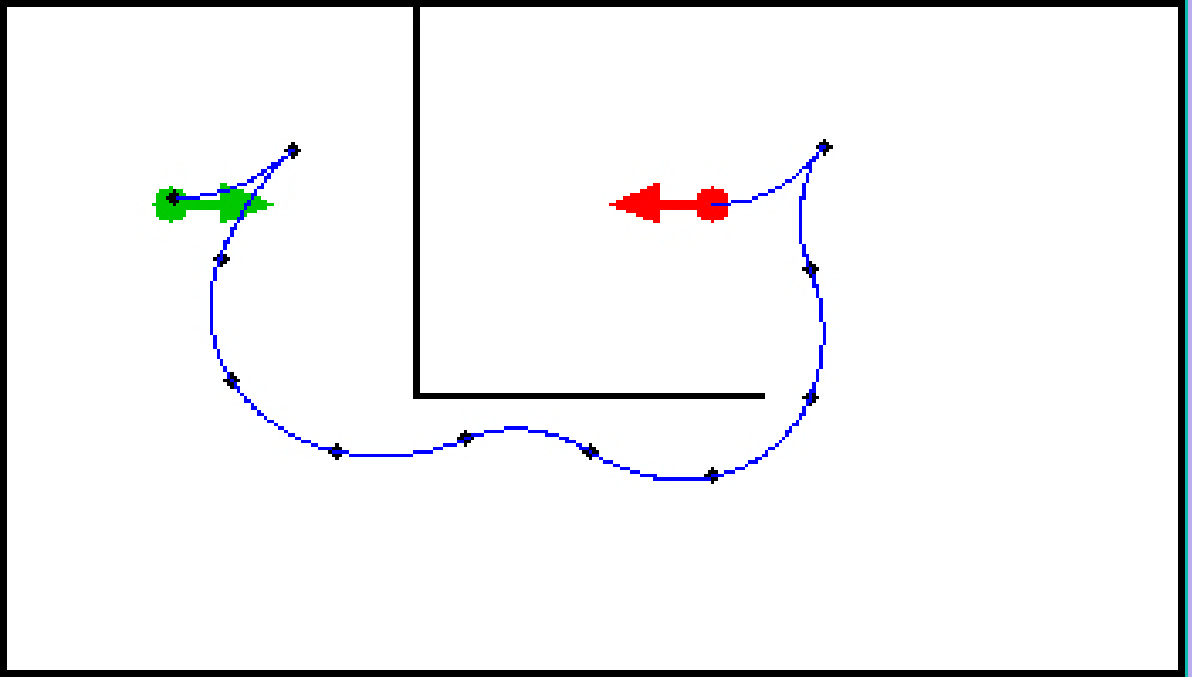}}
  \subfigure[Search tree]{\label{fig:cpa_local_tree}\includegraphics[width=0.24\textwidth]{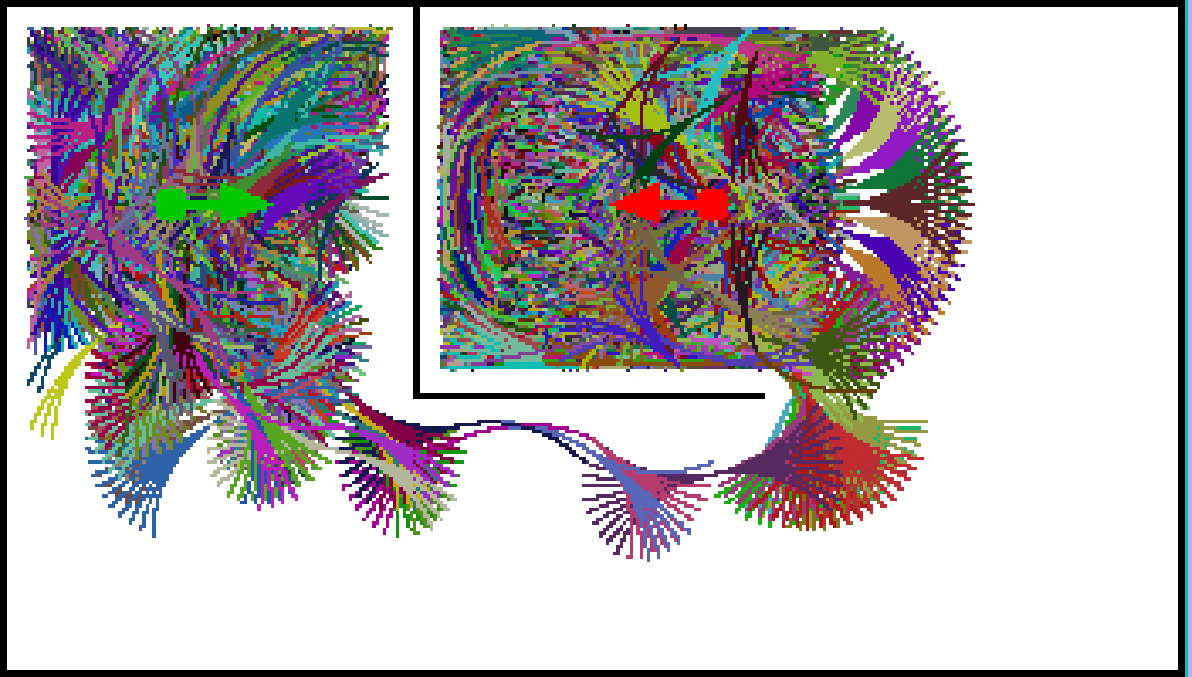}}
  \subfigure[Explored states]{\label{fig:cpa_local_effective}\includegraphics[width=0.24\textwidth]{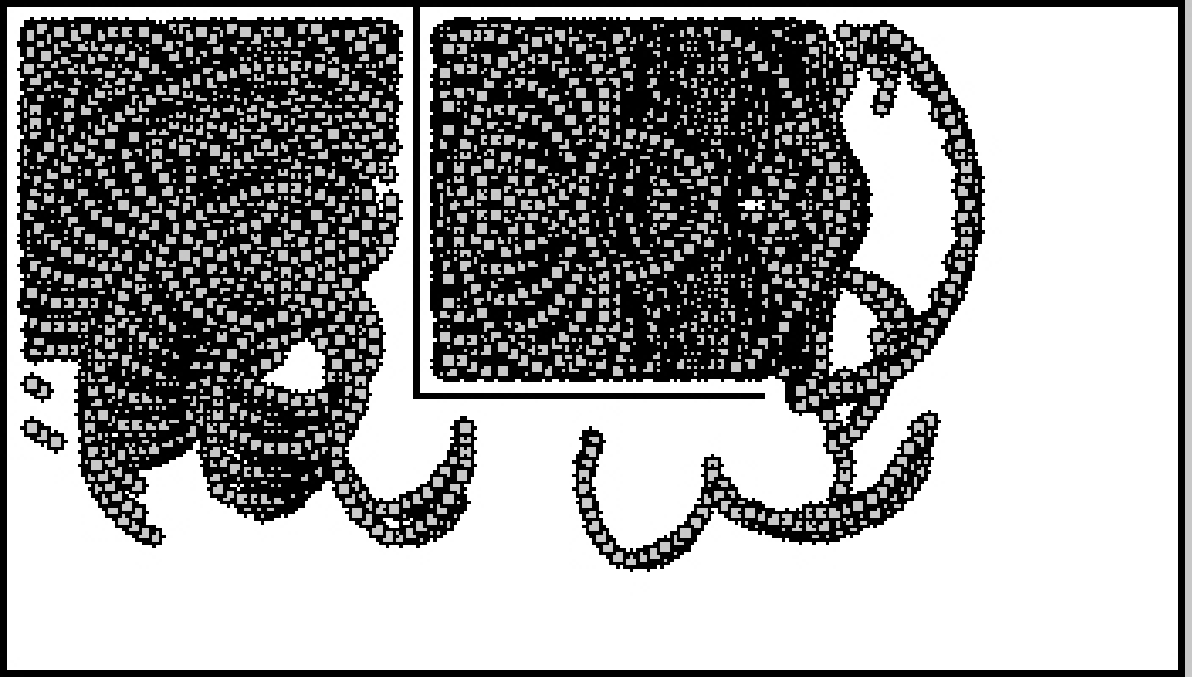}}
  \subfigure[ESD]{\label{fig:cpa_local_distribution}\includegraphics[width=0.24\textwidth]{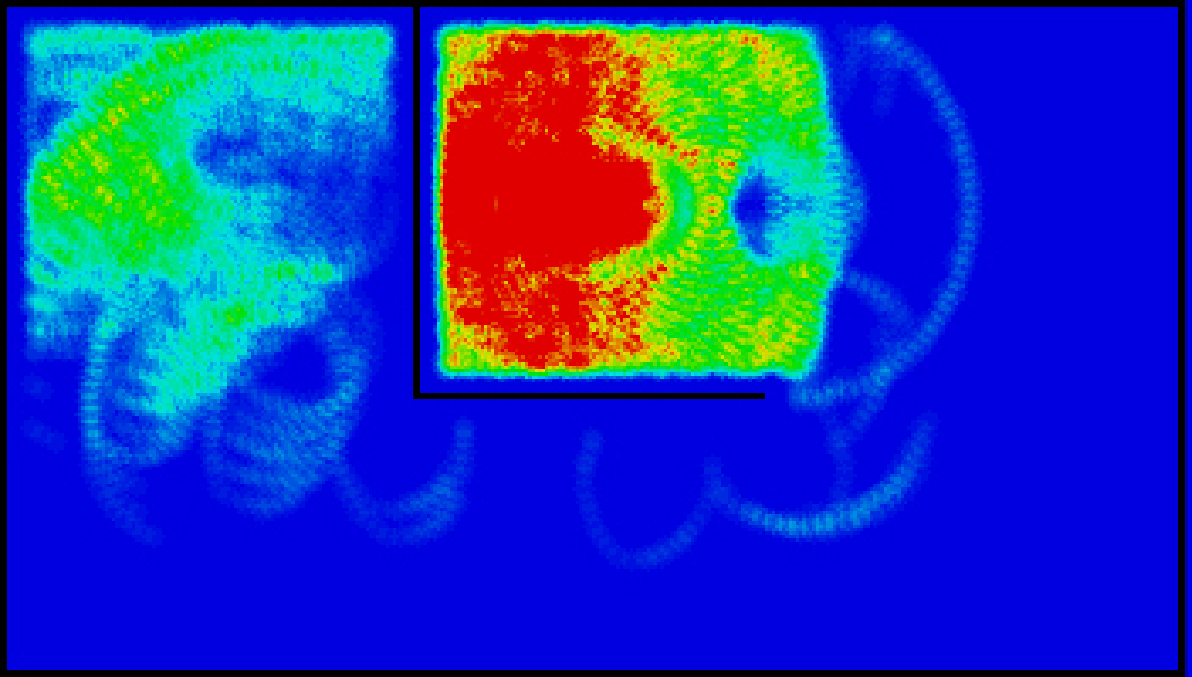}}
  \caption{The Weighted A* that only updates the state of the current location could lead the search to be seriously trapped by the local minimum.}
  \label{fig:cpa_local}
\end{figure*}

The grid map shown in Fig. \ref{fig:cpa_local_path} is a simple environment, where the start and the goal are separated by a deep local minimum. This is a typical situation where the Weighted A* gets trapped. This experiment shows why it is very efficient for the SAS to deal with large clustered environments.

Fig. \ref{fig:cpa_local} is the result of the Weighted A* by inflating the heuristic term with  $\mu=2.0$ in (\ref{eq:anytime}). Fig. \ref{fig:cpa_local_path} shows the produced path of the Weighted A*, the red arrow and the green arrow represent the start and the goal respectively, and the black dots along the path show the joints of the motion primitives. Fig. \ref{fig:cpa_local_tree} shows the search tree of the Weighted A*. As mentioned, the search tree is iteratively constructed by applying a set of predefined motion primitives of a fixed size. The little cells in Fig. \ref{fig:cpa_local_effective} represent the locations of the explored states during the search. Fig. \ref{fig:cpa_local_distribution} shows the explored states distribution (ESD). The ESD shows how many times a certain spot is explored during the search. The red areas represent the most frequently explored areas, whereas the blue areas have not been touched at all.

As in Fig. \ref{fig:cpa_local_distribution}, the area enclosed in the local minimum totally turns red, which clearly shows the search is trapped there. The same effect is also observable in Fig. \ref{fig:cpa_local_tree} and Fig. \ref{fig:cpa_local_effective}. This experiment shows that the local minima can severely increase the computation time and memory cost. 

\begin{figure*}[htbp]
  \centering
  \subfigure[Produced path]{\label{fig:sas_local_path}\includegraphics[width=0.24\textwidth]{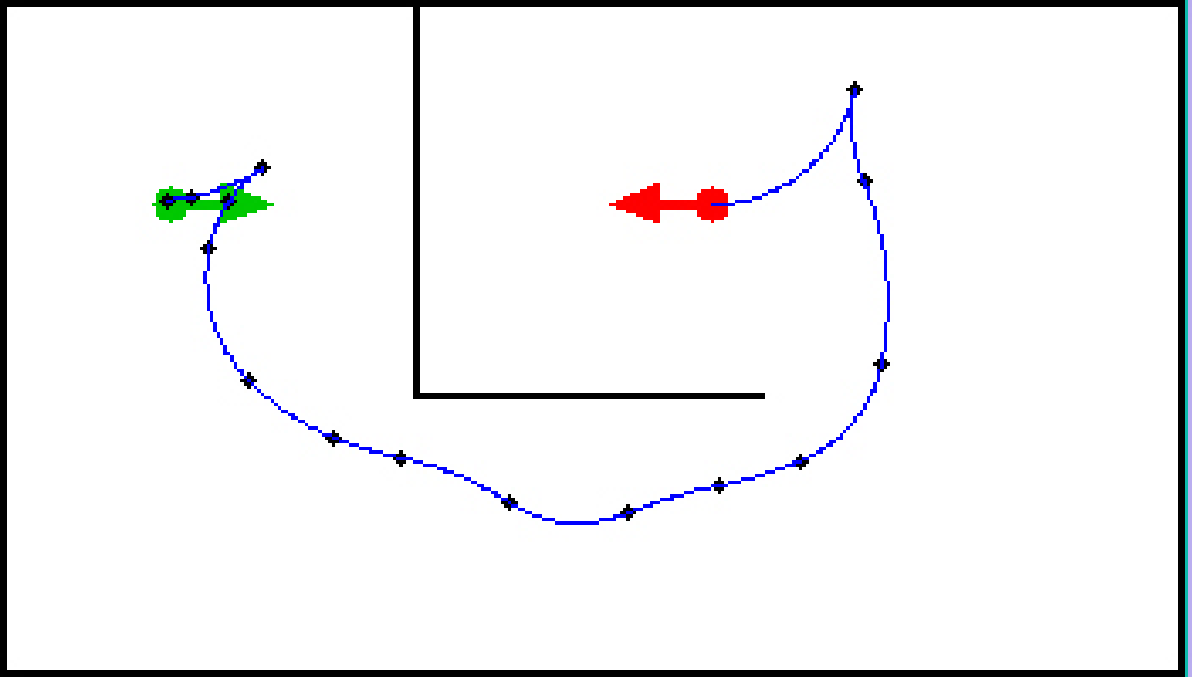}}
  \subfigure[Search tree]{\label{fig:sas_local_tree}\includegraphics[width=0.24\textwidth]{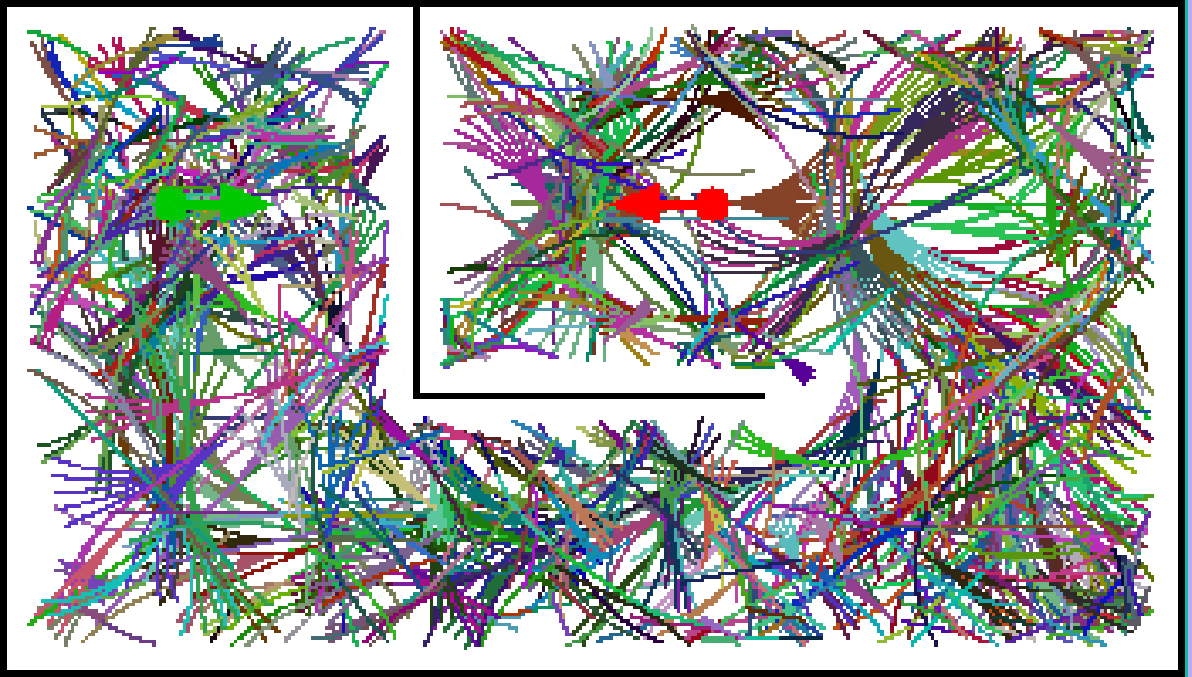}}
  \subfigure[Effective zone]{\label{fig:sas_local_effective}\includegraphics[width=0.24\textwidth]{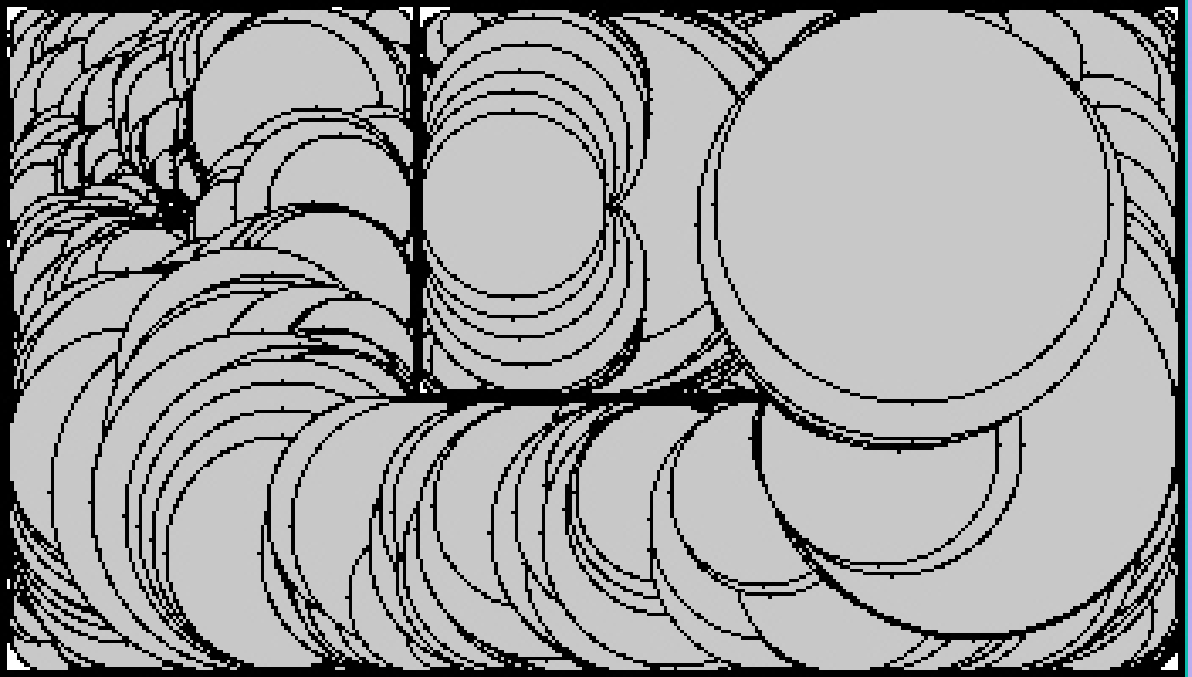}}
  \subfigure[ESD]{\label{fig:sas_local_distribution}\includegraphics[width=0.24\textwidth]{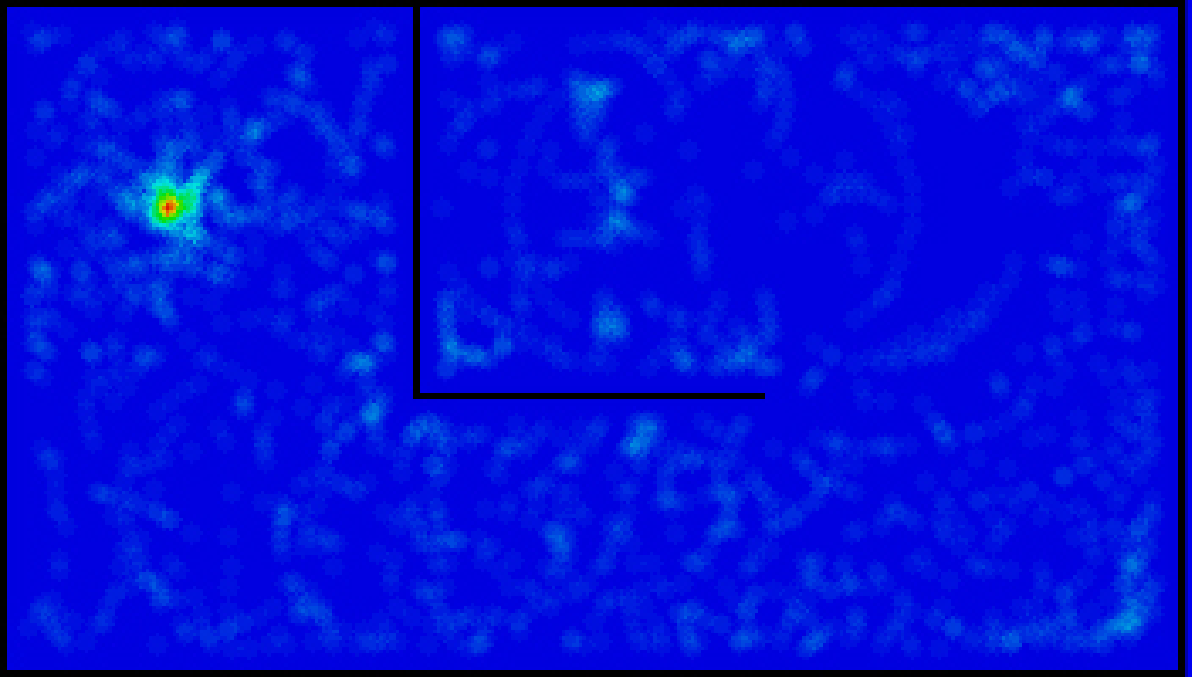}}
  \caption{The SAS can rapidly explore the local minimum and therefore largely reduces the computation time of path planning.}
  \label{fig:sas_local}
\end{figure*}

Fig. \ref{fig:sas_local} is the result of the SAS with the setting of $\kappa^o=1$, $\kappa^g=0.6$ and the minimum step $\lambda=3$ in (\ref{eq:effective_radius_final}). As in Fig. \ref{fig:sas_local_distribution}, only the very small area around the goal is slightly explored and the rest of the space is only sparsely searched, which shows that the SAS has found a path with a very small computation cost. This could also be proved with Table \ref{tab:local}. Compared with the Weighted A*, the computation time and memory cost of the SAS are both largely reduced, but with almost the same path cost. The reason can be explained with Fig. \ref{fig:sas_local_effective}. The circular areas in Fig. \ref{fig:sas_local_effective} represent the effective zones for each explored state. Unlike the Weighted A*, which only updates the states at the current location when exploring a new state [Fig. \ref{fig:cpa_local_effective}], the SAS updates all states within the effective zone. This makes it able to immediately explore the local minimum fully and get out of there right away. For clustered, large-scale environments, the strength of the SAS over the Weighted A* will be more dominant. It is also observable from Table \ref{tab:local} that when $\mu$ is increased from 2.0 to 3.0, the path cost, computation time and the memory cost of  Weighted A* algorithm also increase. This shows that  simply inflating the heuristic does not always mean a reduction in the computation cost. With the existence of the deep local minimum, largely inflating the heuristic may also make the Weighted A* algorithm be trapped longer.

\begin{table}[t]
  \caption{Experiment result under the environment of deep local minimum}\label{tab:local}
  \begin{tabular*}{0.48\textwidth}{@{\extracolsep\fill}llll}
    \hline
    & Path Cost & Computation Time 	& Memory Cost\\
    \hline 
    WA*	($\mu=1.0$)			& 500.75    & 8556.76 ms        & 105168\\
    WA*	($\mu=2.0$)			& 500.70    & 1860.74 ms        & 45570\\
    WA*	($\mu=3.0$)			& 539.37    & 3148.71 ms        & 63879\\ 
    SAS 				 	& 549.89    & 72.14 ms          & 2355\\ 
    \hline
  \end{tabular*}
\end{table}

\subsection{Around the Goal}\label{sec:goal}
\begin{figure*}[htbp]
  \centering
  \subfigure[Produced path]{\label{fig:ccs_close_to_goal_path}\includegraphics[width=0.24\textwidth]{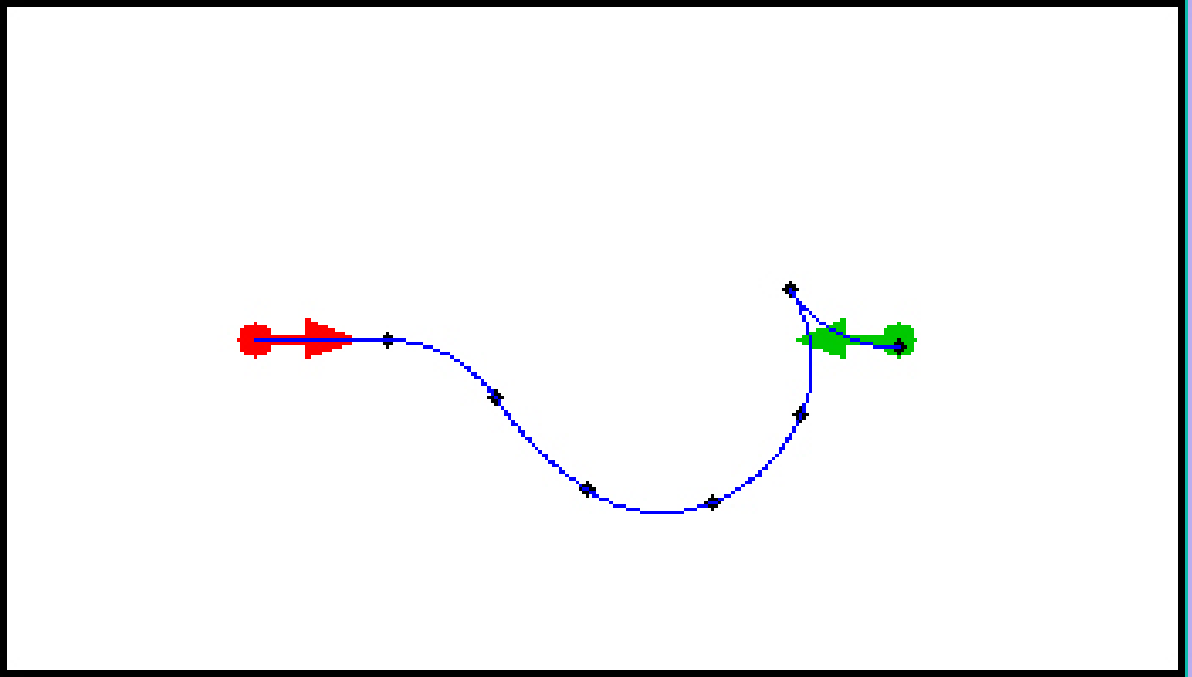}}
  \subfigure[Search tree]{\label{fig:ccs_close_to_goal_tree}\includegraphics[width=0.24\textwidth]{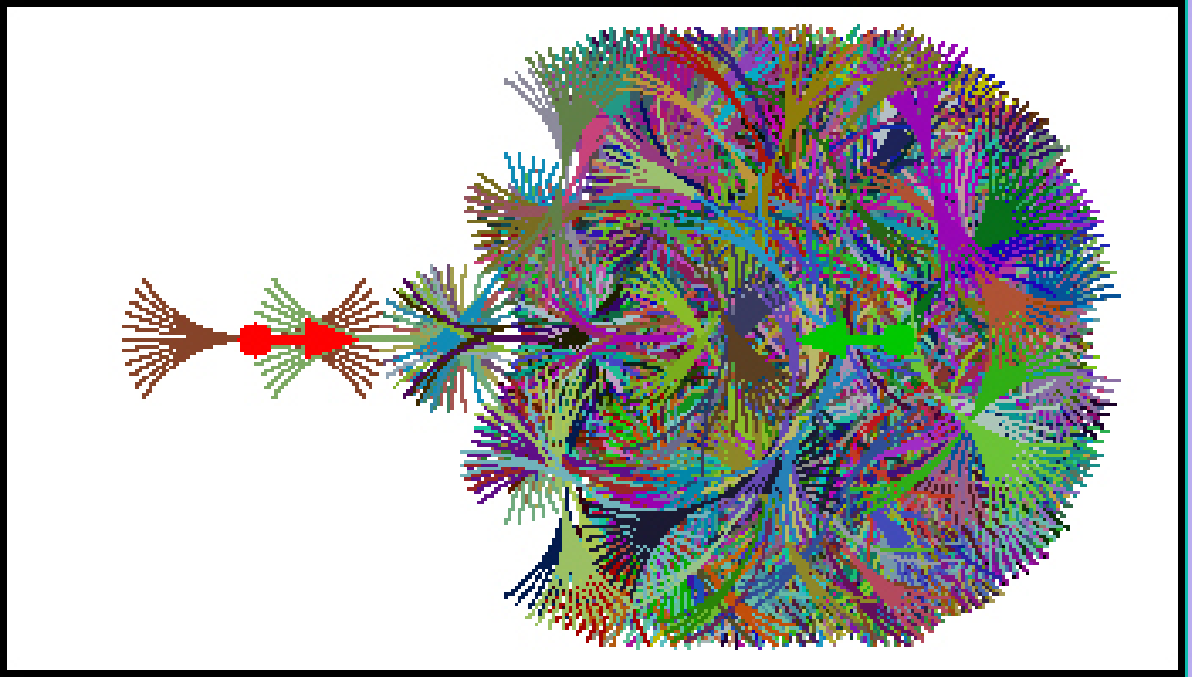}}
  \subfigure[Explored states]{\label{fig:ccs_close_to_goal_effective}\includegraphics[width=0.24\textwidth]{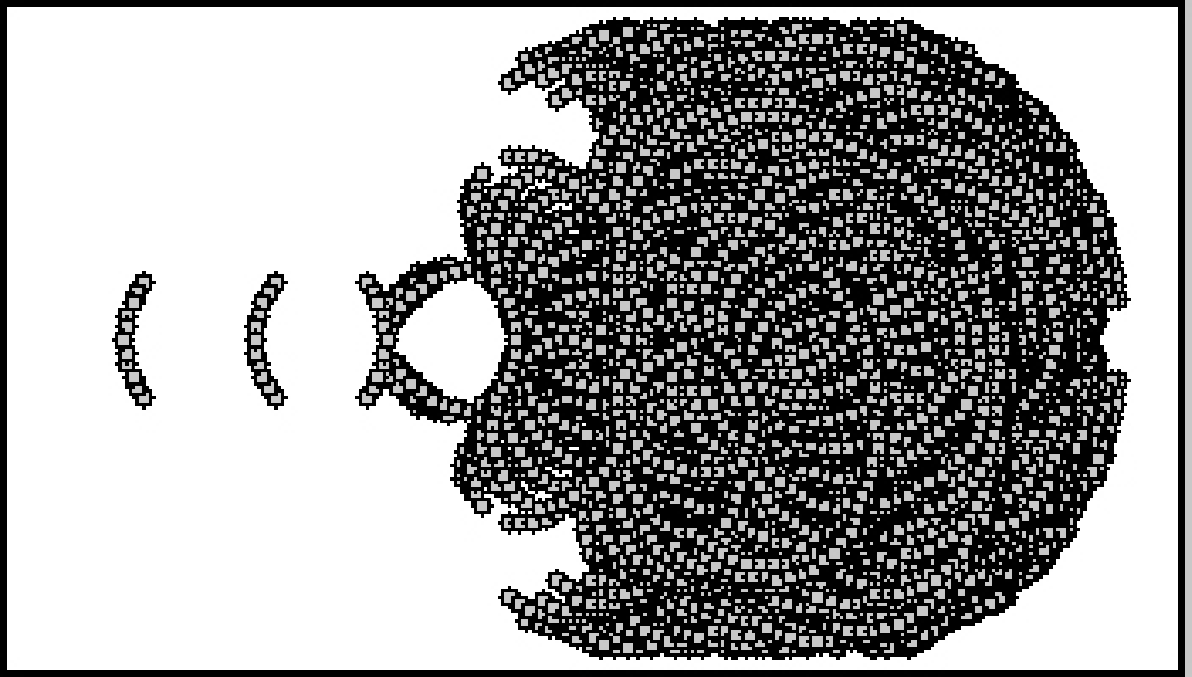}}
  \subfigure[ESD]{\label{fig:ccs_close_to_goal_distribution}\includegraphics[width=0.24\textwidth]{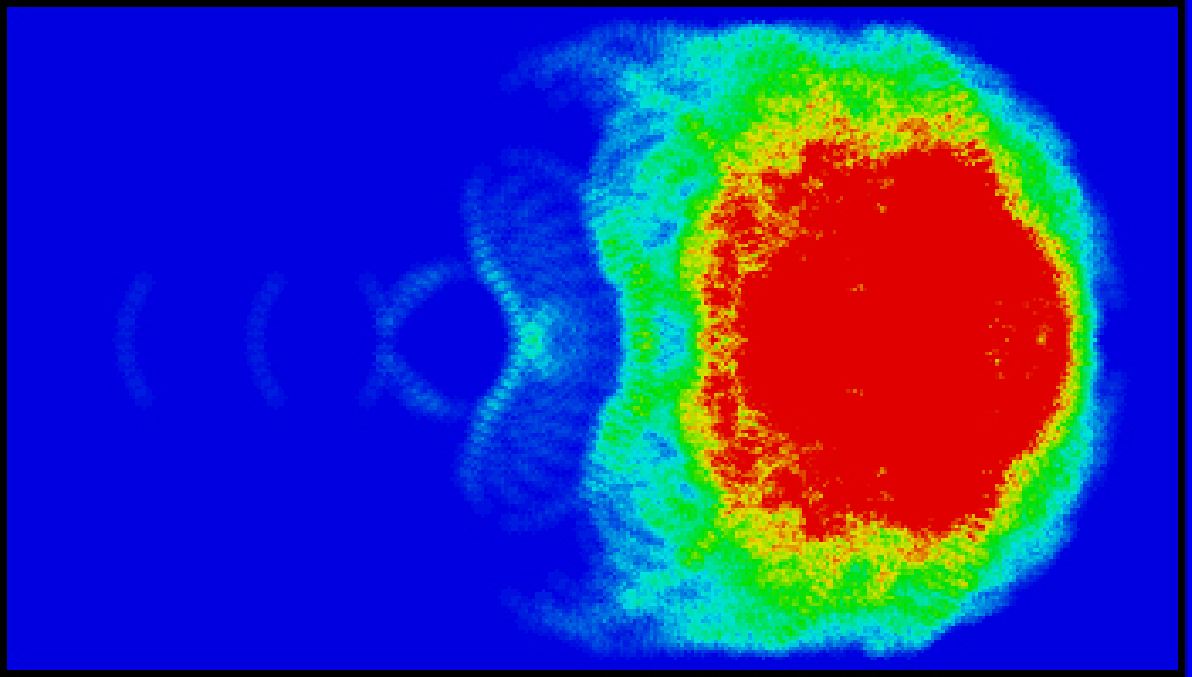}}
  \caption{The Weighted A* ($\mu =2.0$) that explores the state-space with fixed motion primitives makes it very hard to land exactly on the goal state.}
  \label{fig:ccs_close_to_goal}
\end{figure*}

This experiment shows the effect of $\kappa^g$ of the SAS, while the search approaches the goal by increasingly setting $\kappa^g$ from 0.1 to 1.0. The other parameters are set constant ($\kappa^o=1$ and $\lambda=3$).  
The grid map shown in Fig. \ref{fig:ccs_close_to_goal_path} is a small environment where the start and the goal are set very close to each other. 

Fig. \ref{fig:ccs_close_to_goal} is the result of the Weighted A* ($\mu =2.0$). Fig. \ref{fig:ccs_close_to_goal_distribution} clearly shows that the area around the goal turns totally red, which means it has been very intensively explored during the search. This is also observable from Fig. \ref{fig:ccs_close_to_goal_tree} and Fig. \ref{fig:ccs_close_to_goal_effective}.

\begin{figure*}[htbp]
  \centering
  \subfigure[Produced path]{\label{fig:sas_close_to_goal_path_0.2}\includegraphics[width=0.24\textwidth]{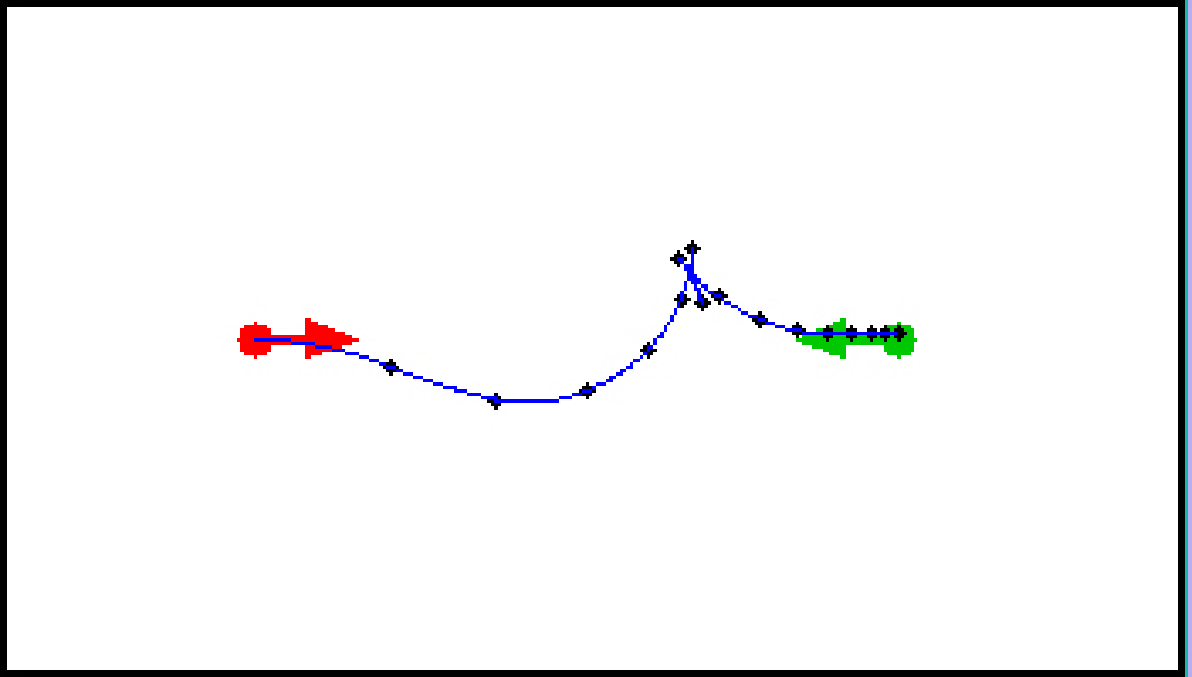}}
  \subfigure[Search tree]{\label{fig:sas_close_to_goal_tree_0.2}\includegraphics[width=0.24\textwidth]{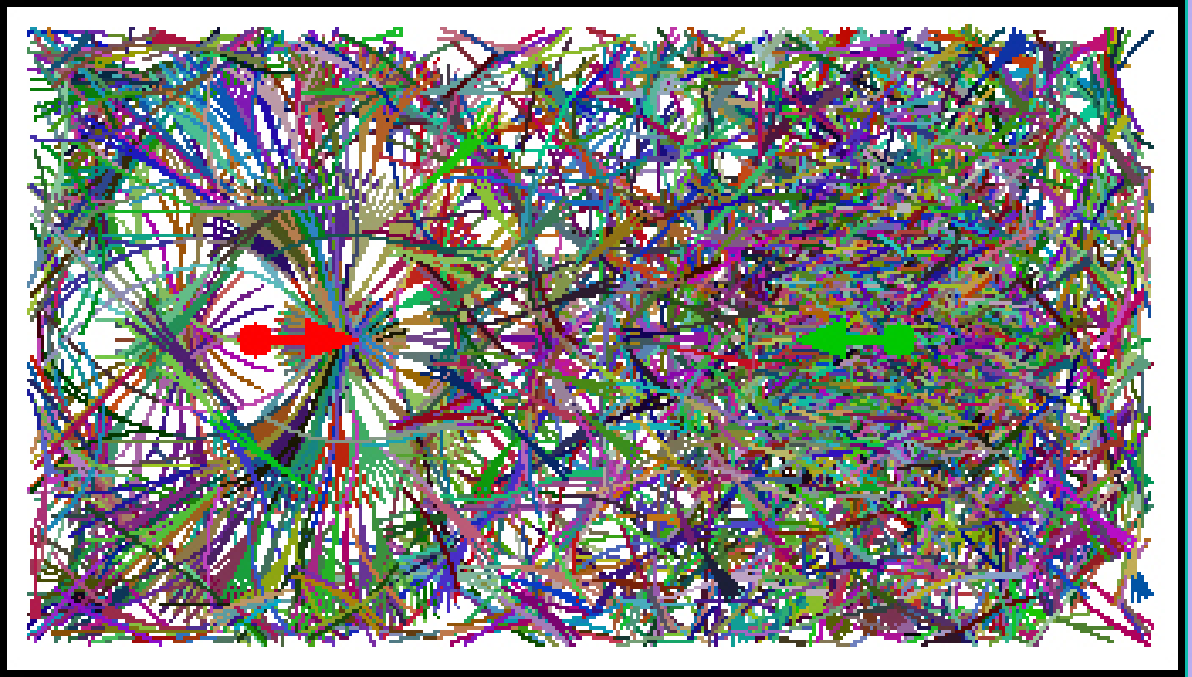}}
  \subfigure[Effective zone]{\label{fig:sas_close_to_goal_effective_0.2}\includegraphics[width=0.24\textwidth]{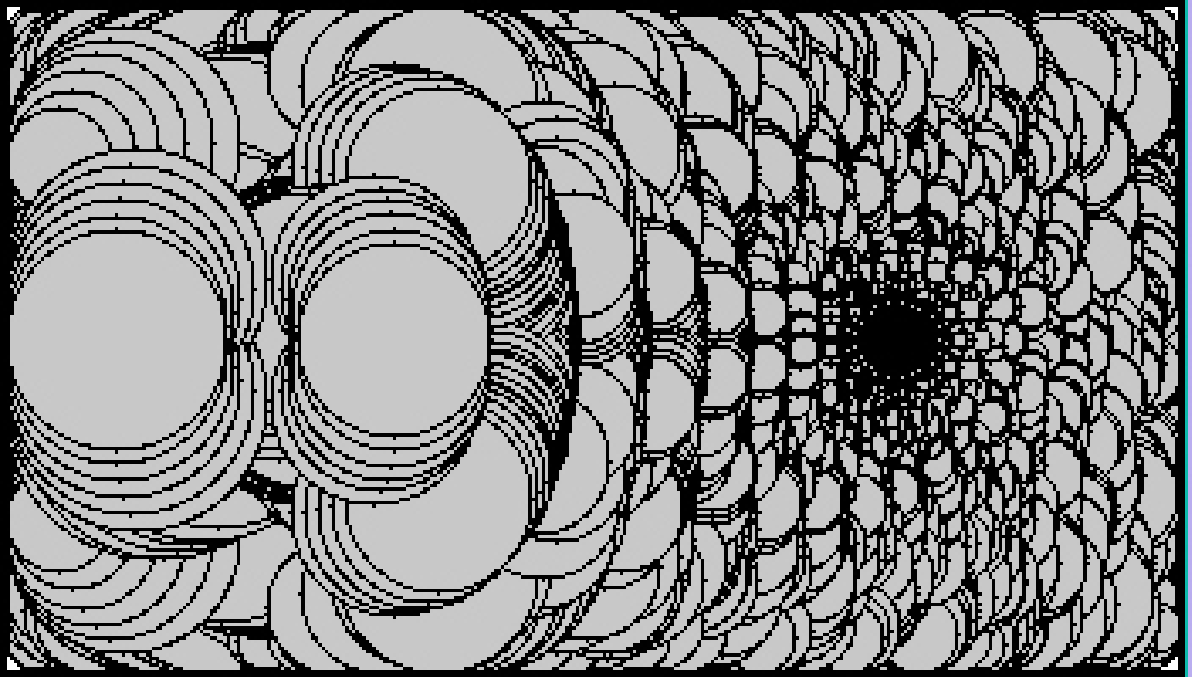}}
  \subfigure[ESD]{\label{fig:sas_close_to_goal_distribution_0.2}\includegraphics[width=0.24\textwidth]{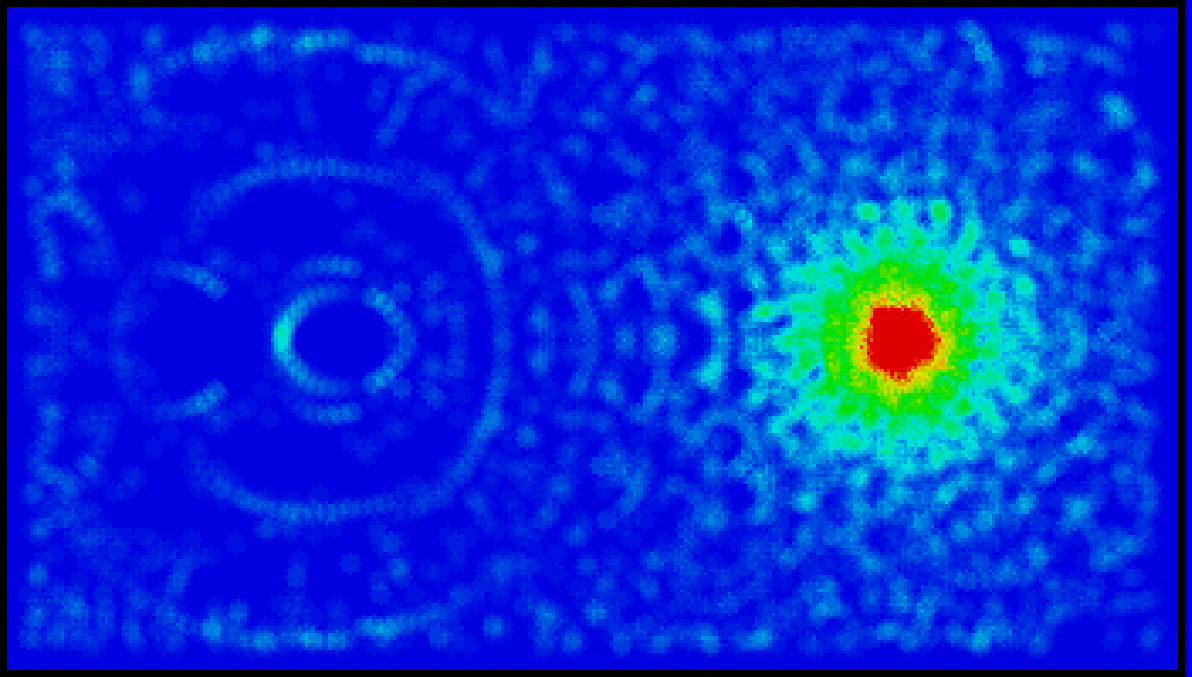}}

  \subfigure[Produced path]{\label{fig:sas_close_to_goal_path_0.5}\includegraphics[width=0.24\textwidth]{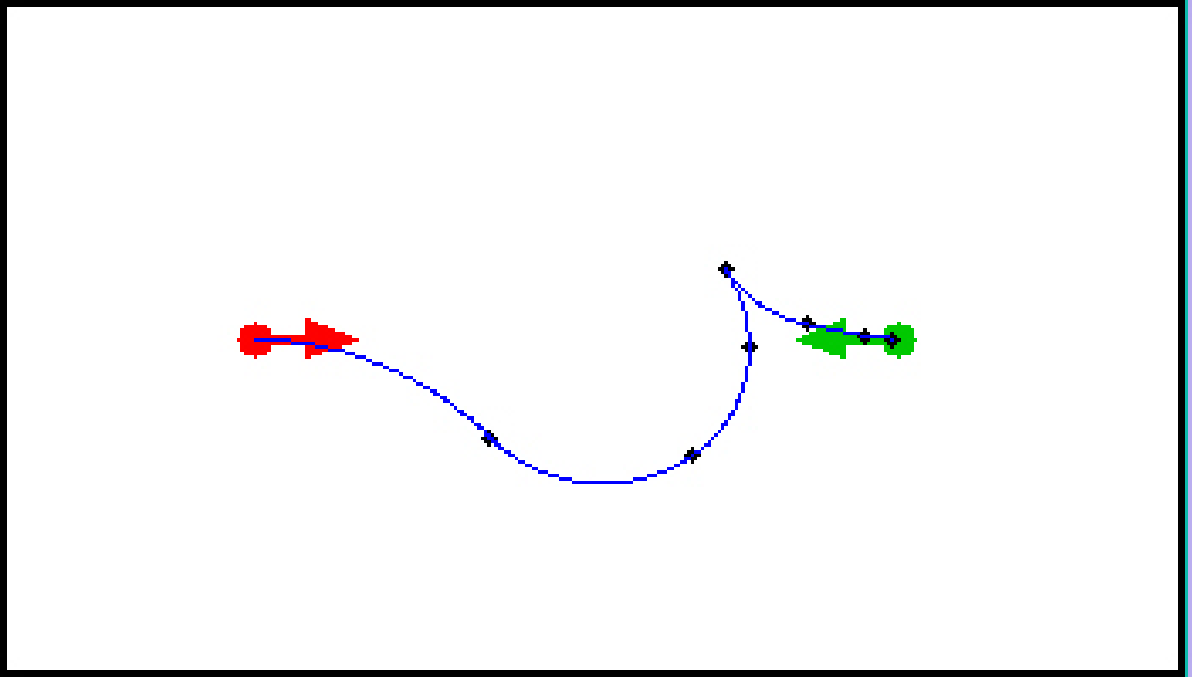}}
  \subfigure[Search tree]{\label{fig:sas_close_to_goal_tree_0.5}\includegraphics[width=0.24\textwidth]{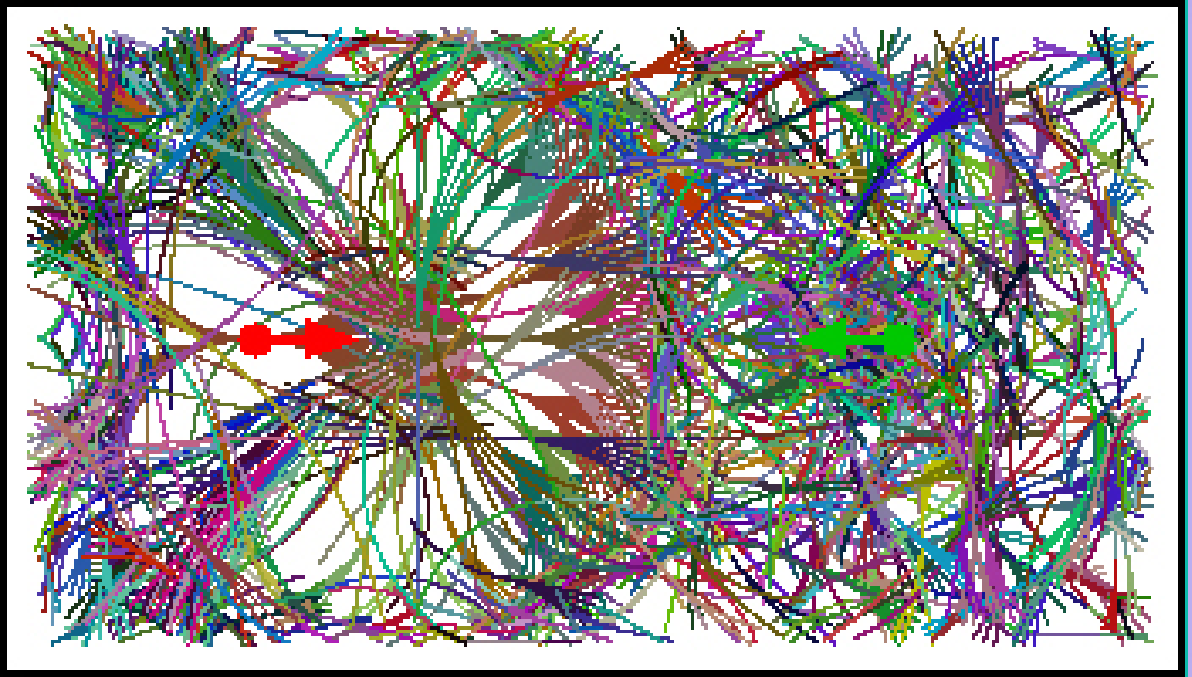}}
  \subfigure[Effective zone]{\label{fig:sas_close_to_goal_effective_0.5}\includegraphics[width=0.24\textwidth]{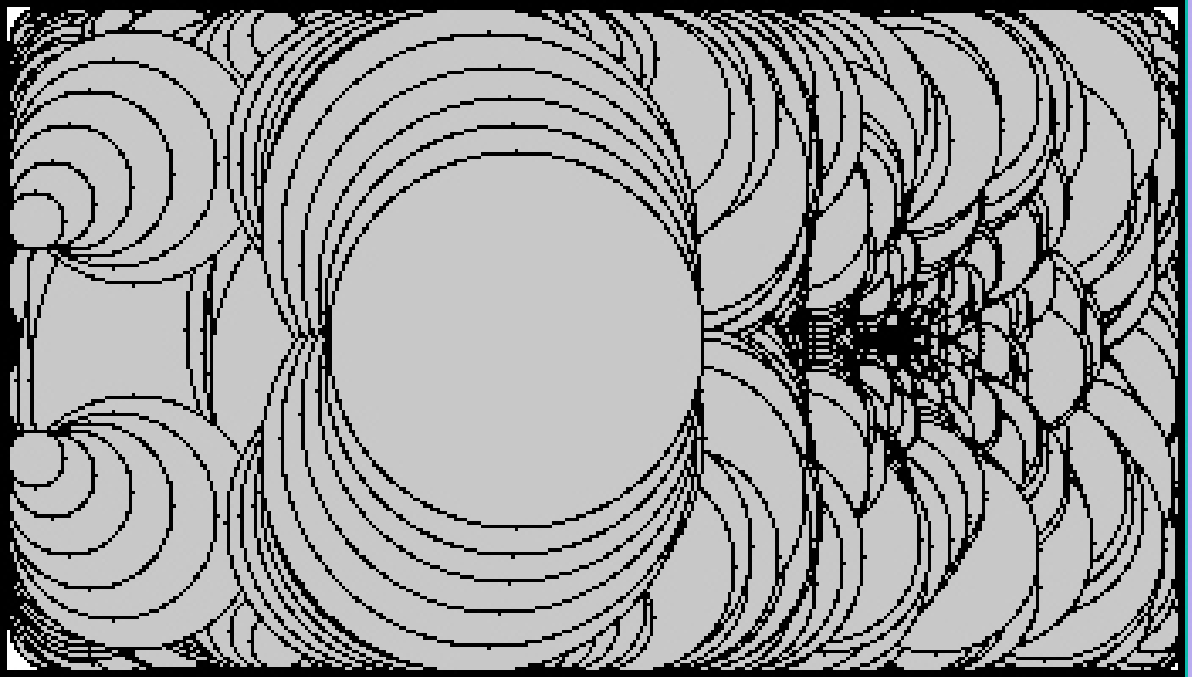}}
  \subfigure[ESD]{\label{fig:sas_close_to_goal_distribution_0.5}\includegraphics[width=0.24\textwidth]{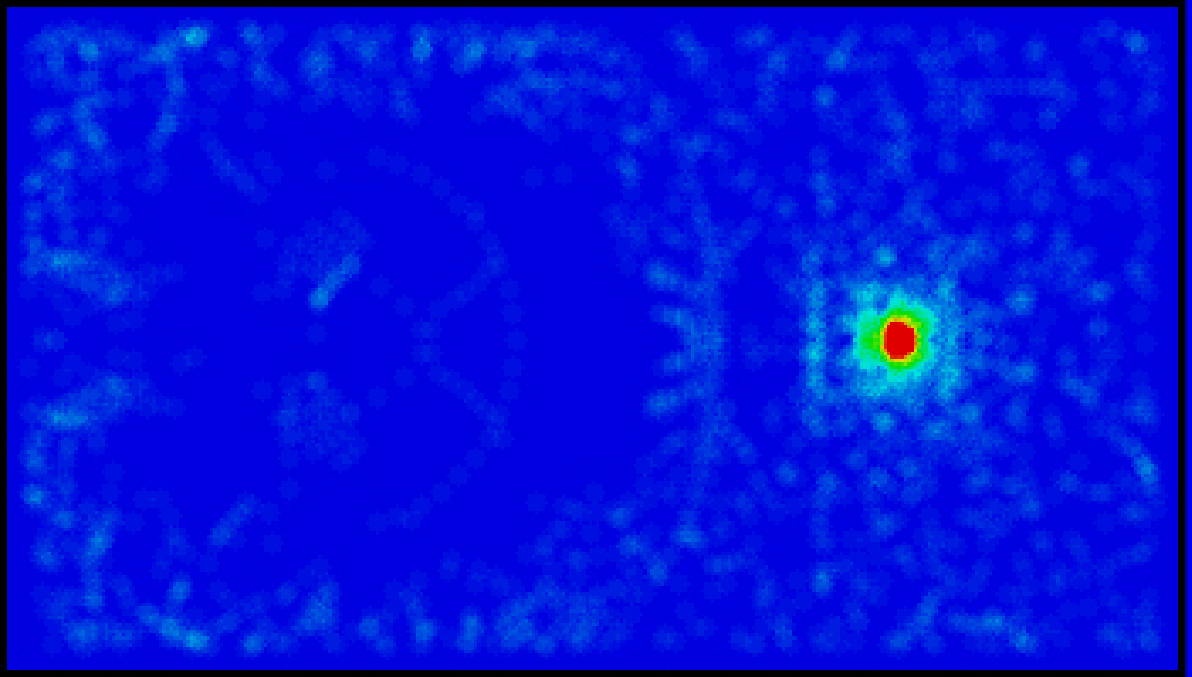}}

  \subfigure[Produced path]{\label{fig:sas_close_to_goal_path_0.8}\includegraphics[width=0.24\textwidth]{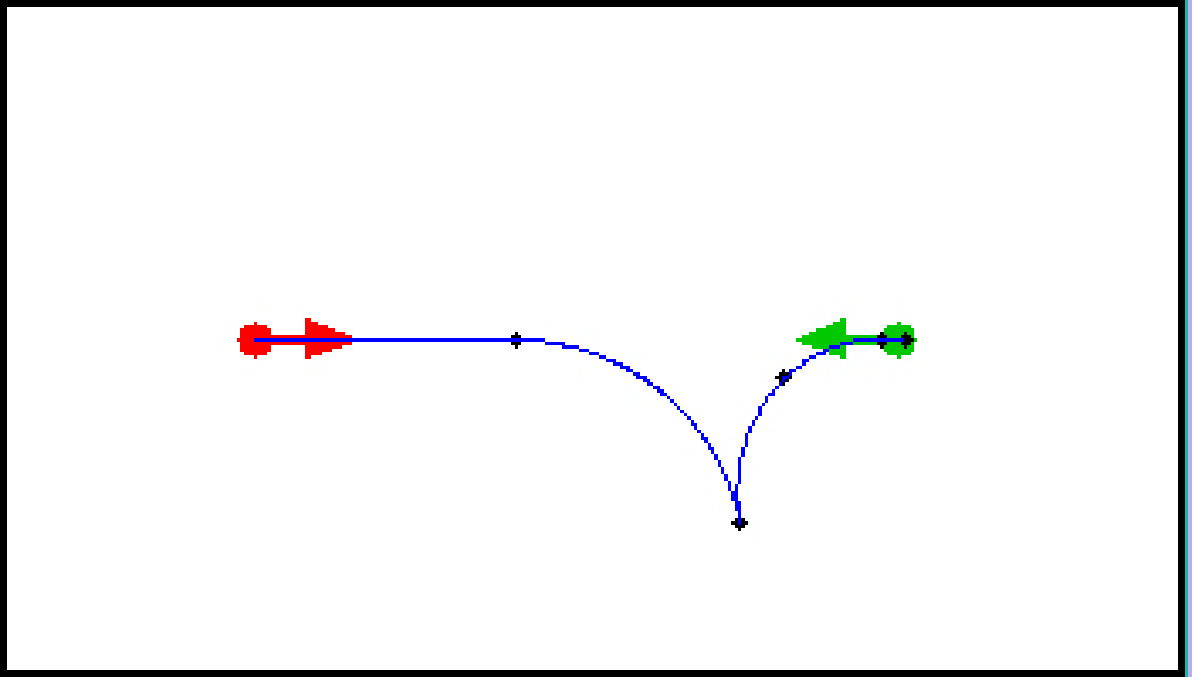}}
  \subfigure[Search tree]{\label{fig:sas_close_to_goal_tree_0.8}\includegraphics[width=0.24\textwidth]{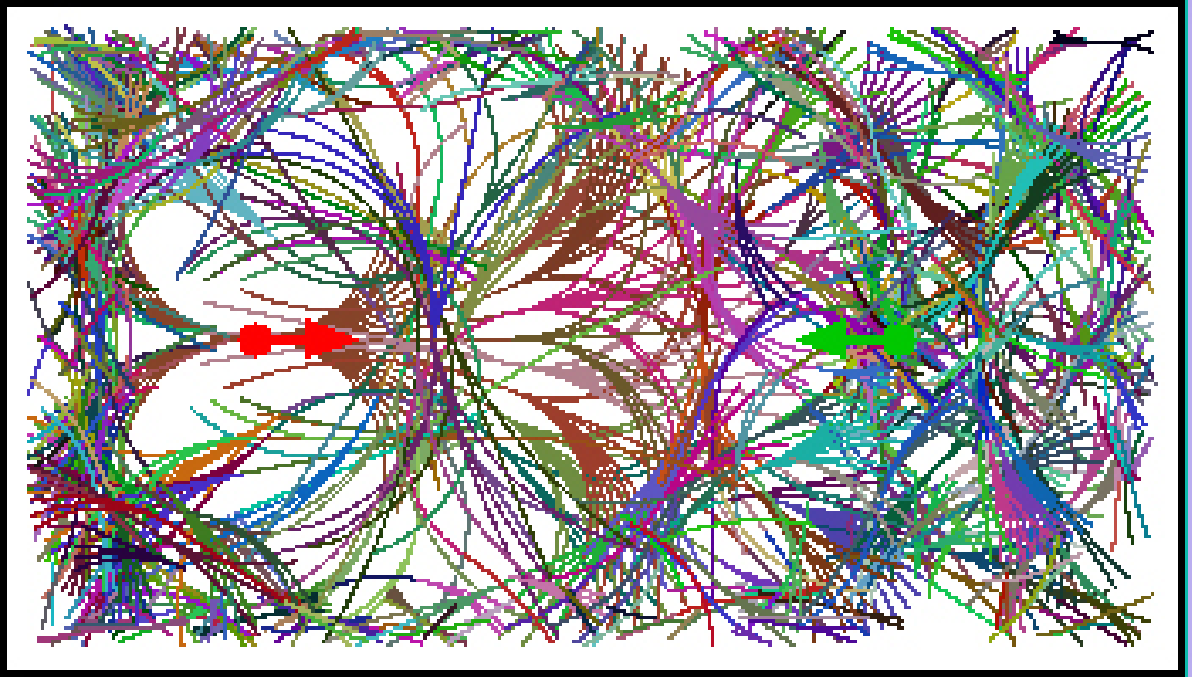}}
  \subfigure[Effective zone]{\label{fig:sas_close_to_goal_effective_0.8}\includegraphics[width=0.24\textwidth]{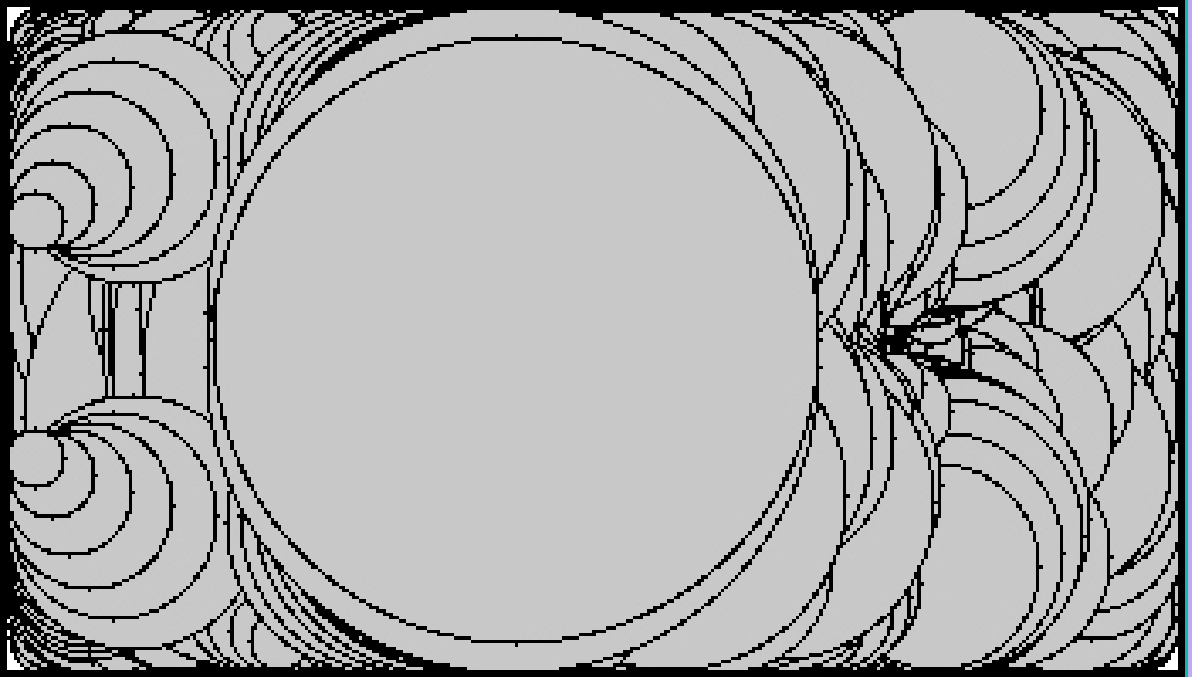}}
  \subfigure[ESD]{\label{fig:sas_close_to_goal_distribution_0.8}\includegraphics[width=0.24\textwidth]{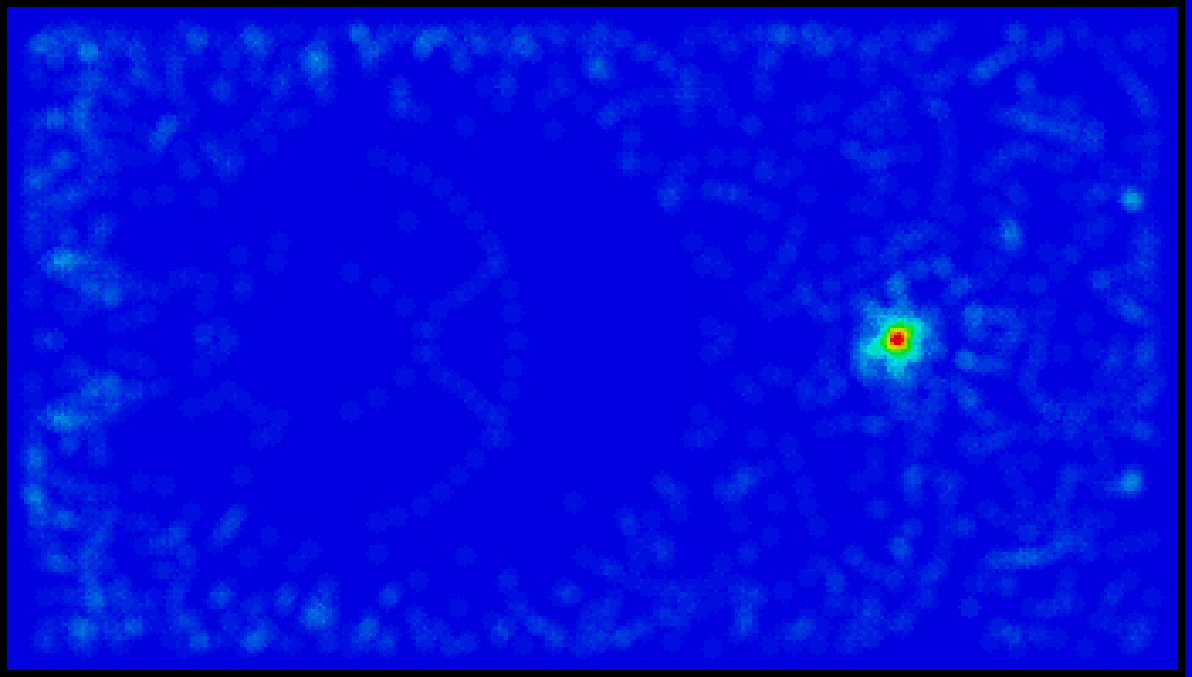}}
  \caption{The search of the SAS converges to the goal faster while $\kappa^g$ increases. (a)$\sim$(d) are the results of the SAS by setting $\kappa^g=0.2$. (e)$\sim$(h) are the results of the SAS by setting $\kappa^g=0.5$. (i)$\sim$(l) are the results of the SAS by setting $\kappa^g=0.8$.}
  \label{fig:sas_close_to_goal_0.8}
\end{figure*}

Fig. \ref{fig:sas_close_to_goal_path_0.2}$\sim$\ref{fig:sas_close_to_goal_distribution_0.2} show the result of SAS by setting $\kappa^g=0.2$. The circular areas in Fig. \ref{fig:sas_close_to_goal_effective_0.2} are the effective zones of all the explored states, and it clearly shows the size of the effective zone reduces as the search approaches the goal. Since there are no obstacles around the goal, the effect of $d^g_i$ becomes more dominant over $d^g_o$, according to (\ref{eq:effective_radius_final}). The motion primitives will also be scaled by the factor $\eta_i$ in  (\ref{eq:scalar}). It is observable in Fig. \ref{fig:sas_close_to_goal_path_0.2} that the length of the motion primitives slowly reduces when the search approaches the goal. This makes the search much finer around the goal. As a result, the search converges to the goal much faster than the Weighted A*. As shown in Fig. \ref{fig:sas_close_to_goal_distribution_0.2}, only the area very close to the goal is intensively explored, which saves both computation time and memory cost compared with the Weighted A*.

Fig. \ref{fig:sas_close_to_goal_path_0.5}$\sim$\ref{fig:sas_close_to_goal_distribution_0.5} and Fig. \ref{fig:sas_close_to_goal_path_0.8}$\sim$\ref{fig:sas_close_to_goal_distribution_0.8} are the results of the SAS by setting $\kappa^g$ $0.5$ and $0.8$ respectively. Fig. \ref{fig:sas_close_to_goal_distribution_0.5} and Fig. \ref{fig:sas_close_to_goal_effective_0.8} show that as $\kappa^g$ increases, the converging effect becomes more dominant. Most areas are only sparsely explored, and only the tiny spot around the goal slightly turns green. The size of the effective zone is dramatically reduced as in Fig. \ref{fig:sas_close_to_goal_effective_0.5} and Fig. \ref{fig:sas_close_to_goal_effective_0.8}, so the length of the motion primitives varied faster along the path as in Fig. \ref{fig:sas_close_to_goal_path_0.5} and \ref{fig:sas_close_to_goal_path_0.8}. 

\begin{figure*}[htbp]
  \centering
  \subfigure[Path cost]{\label{fig:no_obs_path_cost}\includegraphics[width=0.3\textwidth]{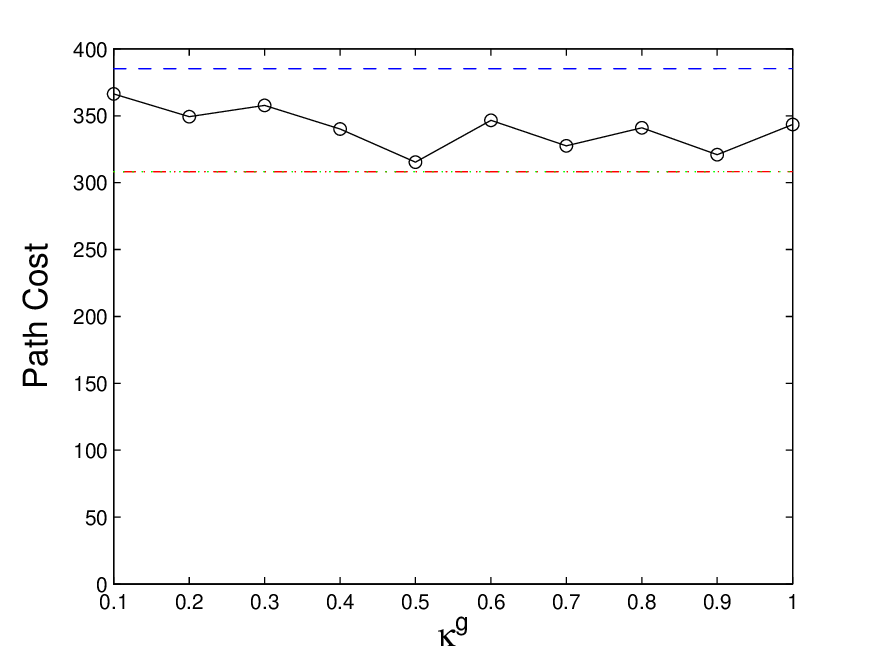}}
  \subfigure[Computation time (ms)]{\label{fig:no_obs_time_cost}\includegraphics[width=0.3\textwidth]{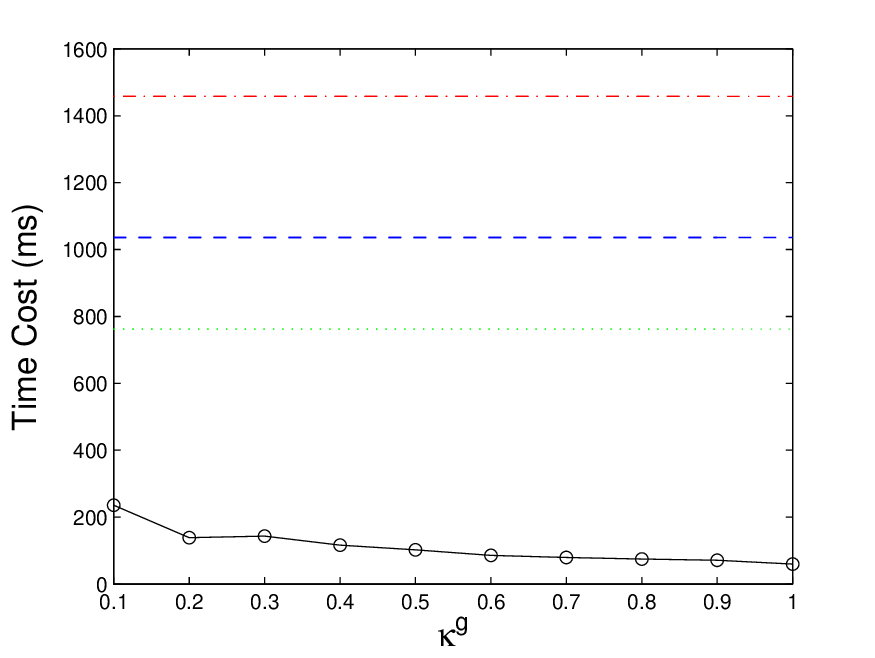}}
  \subfigure[Memory cost]{\label{fig:no_obs_memory_cost}\includegraphics[width=0.3\textwidth]{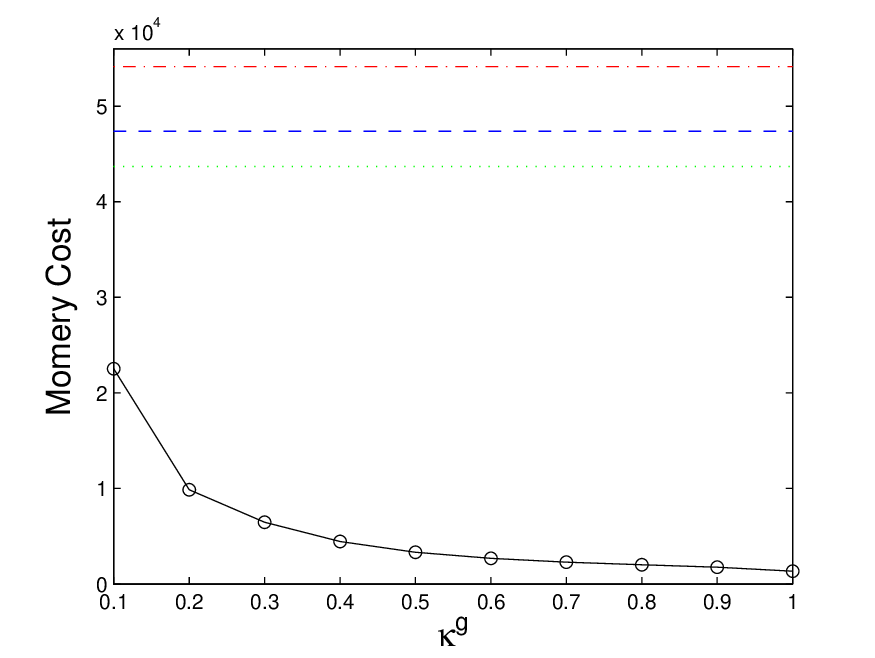}}
  \caption{The black lines in (a)$\sim$(c) show the cost variation of the SAS applying on the grid map in Fig. \ref{fig:sas_close_to_goal_path_0.2} by increasingly setting $\kappa^g$ from $0.1\sim 1.0$.  The colored line shows the result of the Weighted A* with different settings of $\mu$, the red line is the result of setting $\mu=1.0$, the green line and the blue line show the result of setting $\mu=2.0$ and $\mu=3.0$ respectively.}
  \label{fig:no_obs_diagram}
\end{figure*}

Fig. \ref{fig:no_obs_diagram} shows the cost variation of the same experiment. 
The black line shows the result of the SAS, whereas the results of the Weighted A* are shown as the colored lines, the red line represents the results of setting $\mu=1.0$, while the green line and the blue line shows the result of setting $\mu=2.0$ and $\mu=3.0$ respectively. It is observable that the SAS largely reduces the computation cost with almost the same path cost. As shown in Fig. \ref{fig:no_obs_time_cost} and Fig. \ref{fig:no_obs_memory_cost}, the computation cost of the Weighted A* by setting $\mu=3.0$ is higher than the cost by setting $\mu=2.0$, which shows again that inflating heuristic does not always cause a reduction in the computation cost.


The reason that the SAS requires a very low computation cost is not only because of the effective zone, but also due to the scalable motion primitives. The Weighted A* only utilizes the motion primitive of a few predefined lengths, it is very hard for the Weighted A* to end exactly on the goal. This makes the following steps easily go beyond the goal, even when the current step has come very close to the goal. However, the SAS measures the distance $d^g_i$ from the goal to the current state $\zeta_i$ first, then scale the motion primitives to the appropriate length before applying them. This guarantees the search iteratively approaching the goal without going too far, which is another important reason why the SAS converges very fast and requires very low computation cost.

As in Fig. \ref{fig:no_obs_diagram}, the computation time and the memory cost both fell while $\kappa^g$ increased, and the path cost changed only slightly. However, setting $\kappa^g$ with a high value could also reduce the flexibility of the search around goal. Since in Fig. \ref{fig:no_obs_time_cost} and Fig. \ref{fig:no_obs_memory_cost}, the reduction of both the computation time and the memory cost become relatively slow around $\kappa^g=0.6$, setting $\kappa^g$ higher than $0.6$ does not bring many benefits, so $\kappa^g$ should be set around $0.6$.


\begin{figure*}[htbp]
  \centering
  \subfigure[Produced path]{\label{fig:ccs_normal_path}\includegraphics[width=0.4\textwidth]{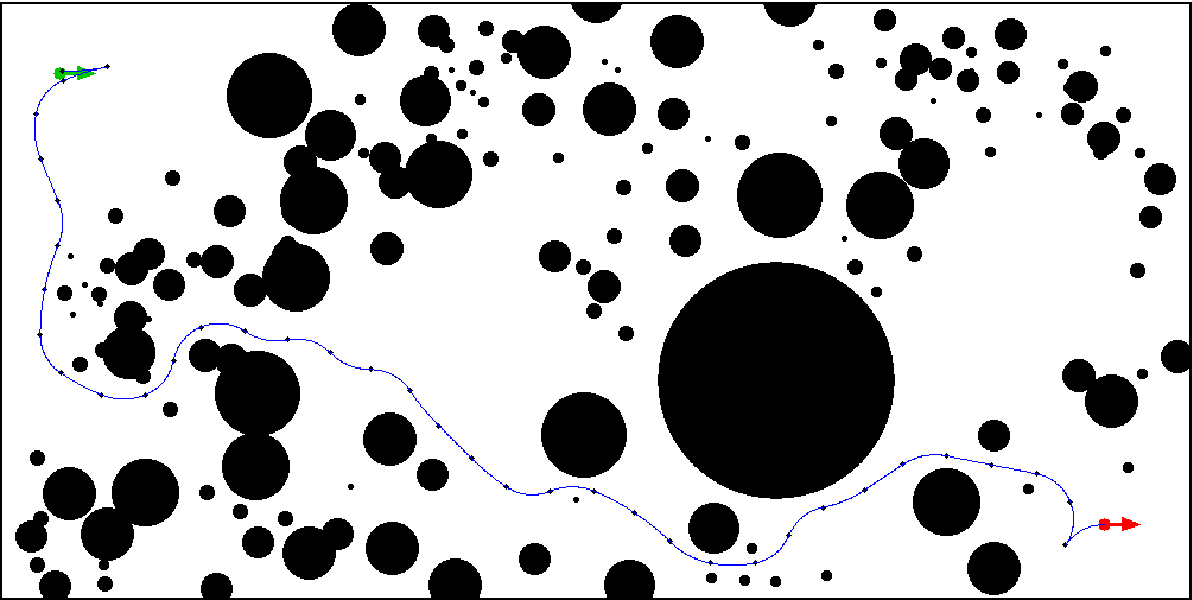}}
  \subfigure[Search tree]{\label{fig:ccs_normal_tree}\includegraphics[width=0.4\textwidth]{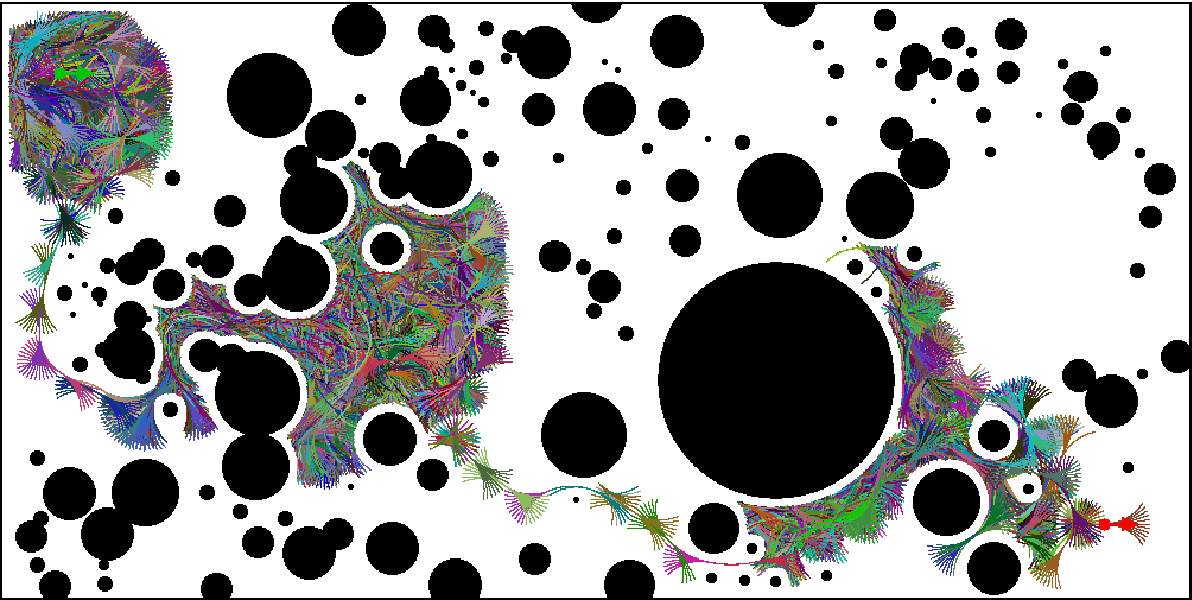}}
  \subfigure[Explored states]{\label{fig:ccs_normal_effective}\includegraphics[width=0.4\textwidth]{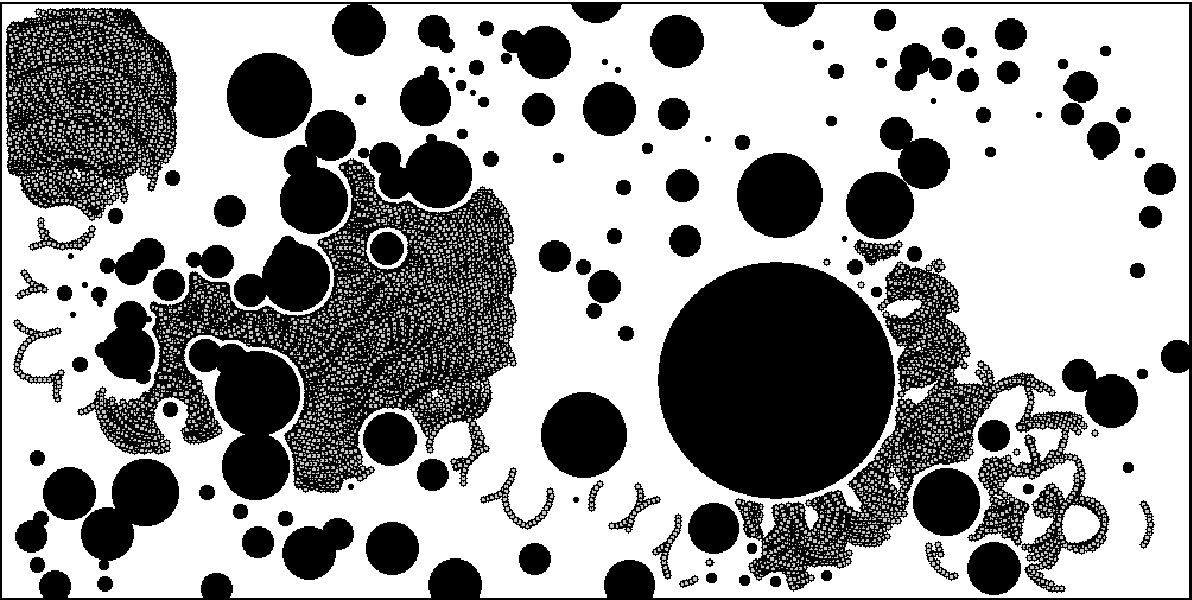}}
  \subfigure[ESD]{\label{fig:ccs_normal_distribution}\includegraphics[width=0.4\textwidth]{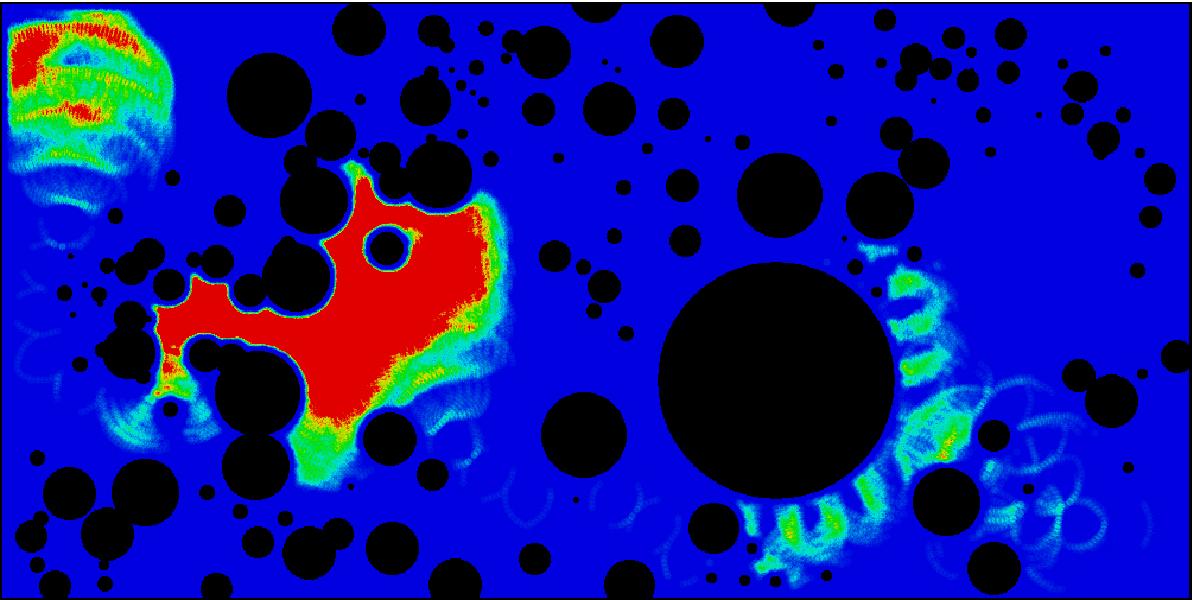}}
  \caption{The result of the Weighted A* by setting $\mu=2.0$. In a clustered environment, the Weighted A* can be seriously trapped by the local minimum, since the states have to be updated one by one. The red zone in (d) shows the location of the local minimum, which is the most intensively explored. The same effect can also be observed in (b) and (c).} 
  \label{fig:ccs_normal}
\end{figure*}

\begin{figure*}[htbp]
  \centering
  \subfigure[Produced path]{\label{fig:sas_normal_0.5_path}\includegraphics[width=0.4\textwidth]{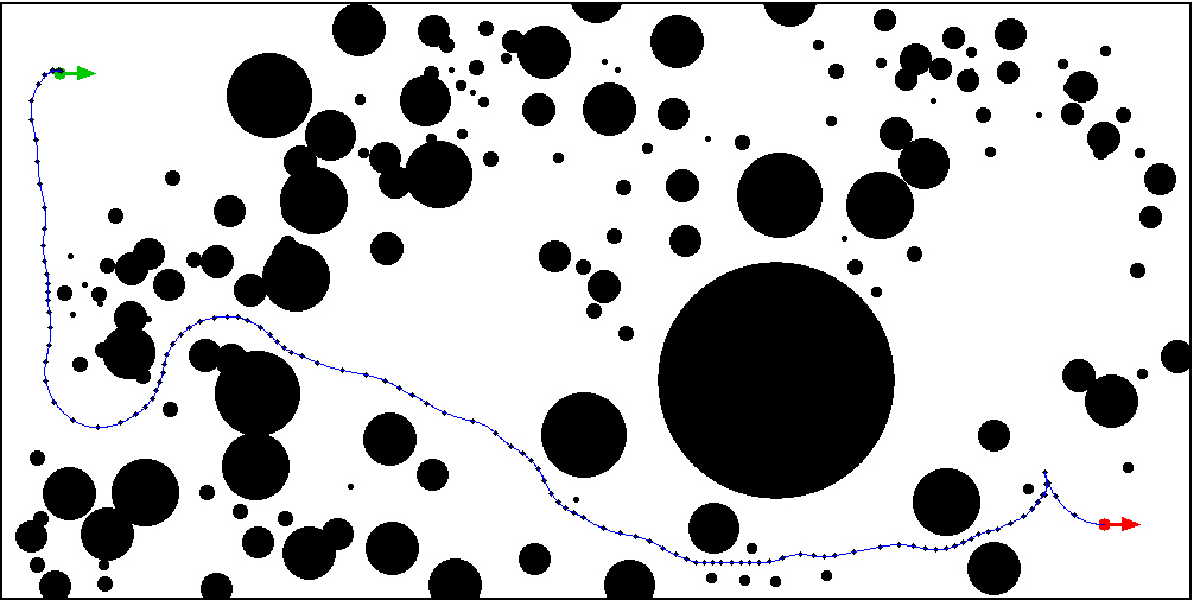}}
  \subfigure[Search tree]{\label{fig:sas_normal_0.5_tree}\includegraphics[width=0.4\textwidth]{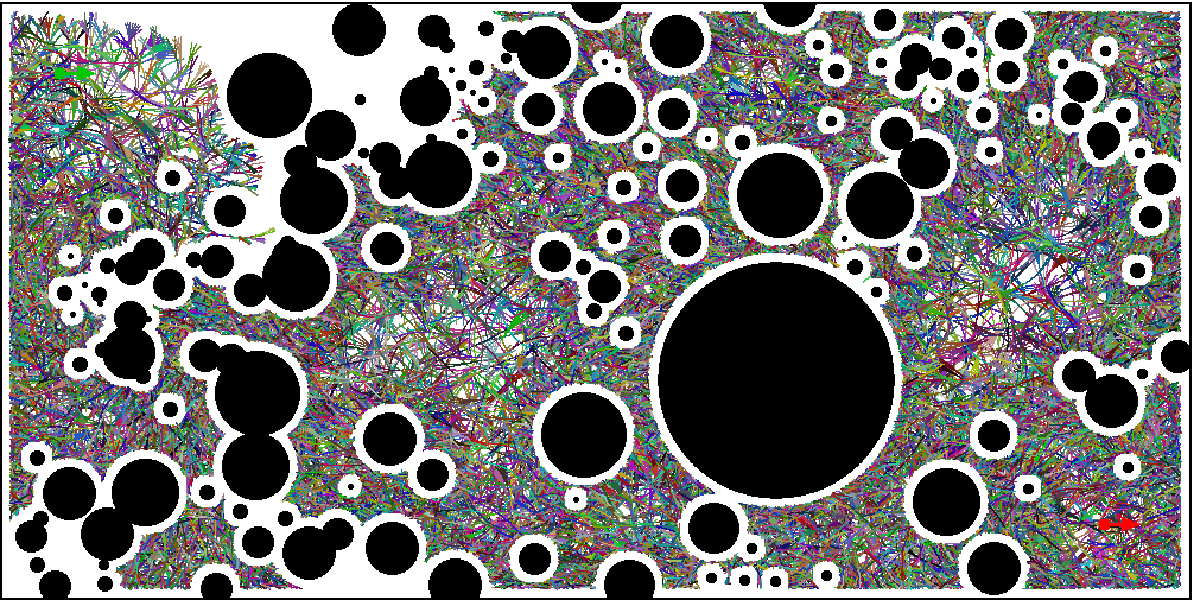}}
  \subfigure[Effective zone]{\label{fig:sas_normal_0.5_effective}\includegraphics[width=0.4\textwidth]{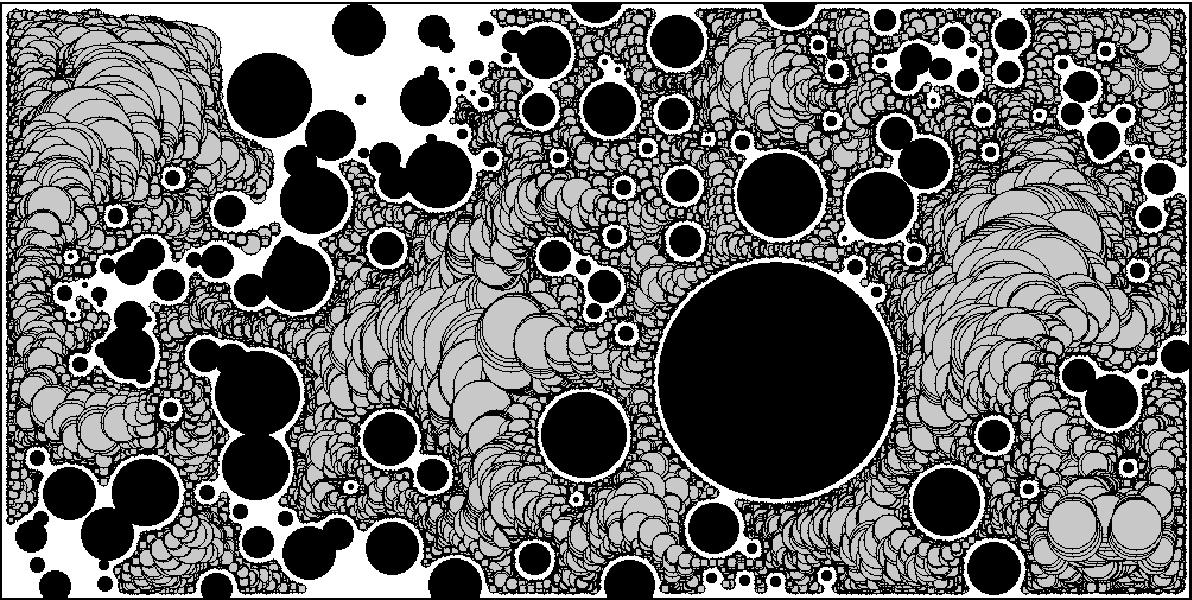}}
  \subfigure[ESD]{\label{fig:sas_normal_0.5_distribution}\includegraphics[width=0.4\textwidth]{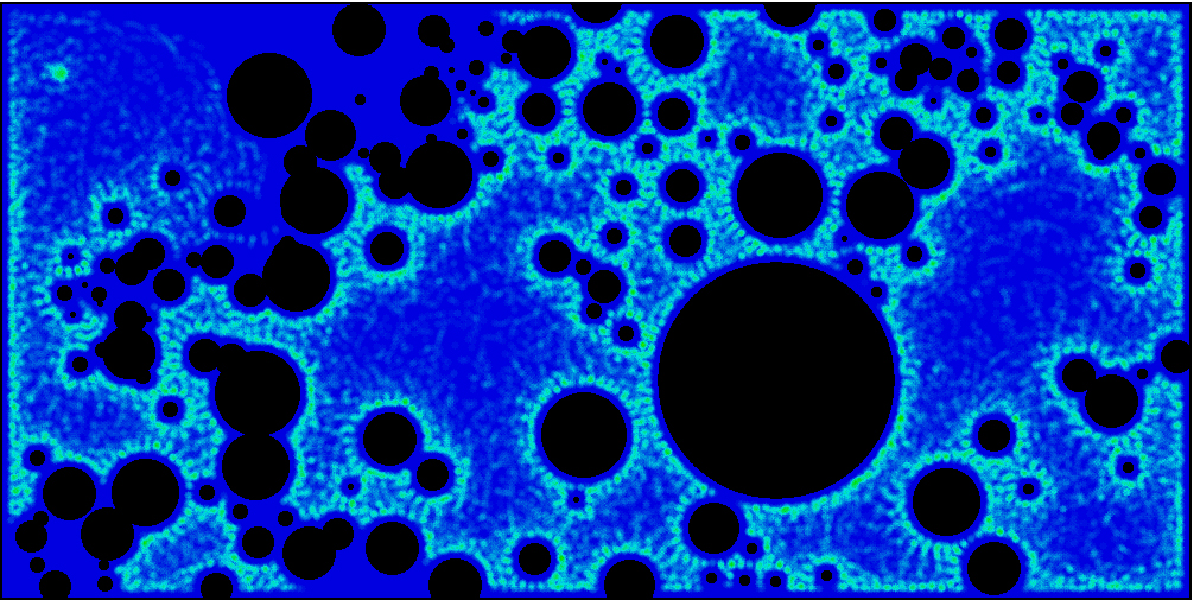}}
  \caption{The result of SAS by setting $\kappa^o=0.5$: (a) and (b) show that the length of the motion primitives will reduce while the search approaches the obstacles. (c) shows how the effective zone expends while the search enters open areas, and shrinks when it approaches the obstacles or the goal. (d) shows that the SAS sparsely explores open areas and intensively explores the space close to obstacles. }
  \label{fig:sas_normal_0.5}
\end{figure*}

\begin{figure*}[htbp]
  \centering
  \subfigure[Produced path]{\label{fig:sas_normal_1_path}\includegraphics[width=0.4\textwidth]{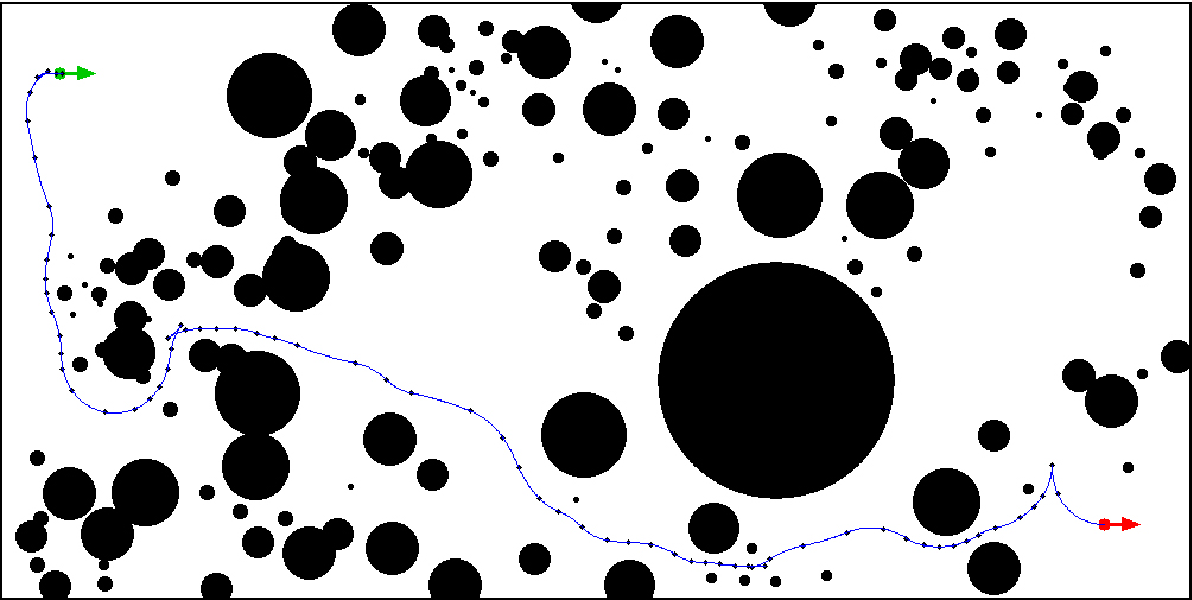}}
  \subfigure[Search tree]{\label{fig:sas_normal_1_tree}\includegraphics[width=0.4\textwidth]{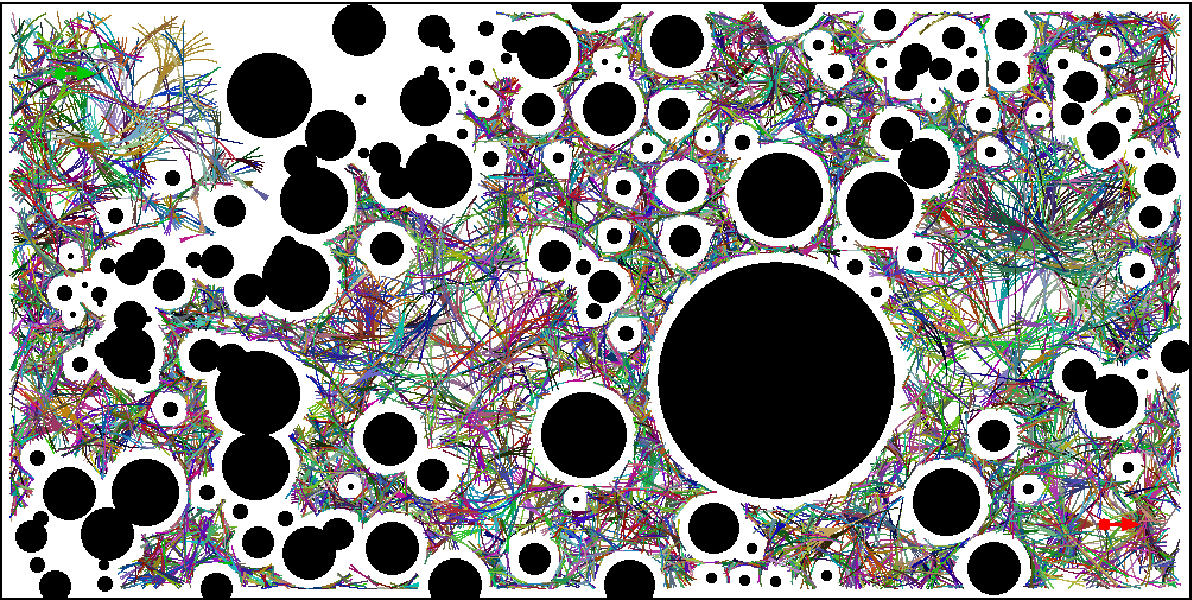}}
  \subfigure[Effective zone]{\label{fig:sas_normal_1_effective}\includegraphics[width=0.4\textwidth]{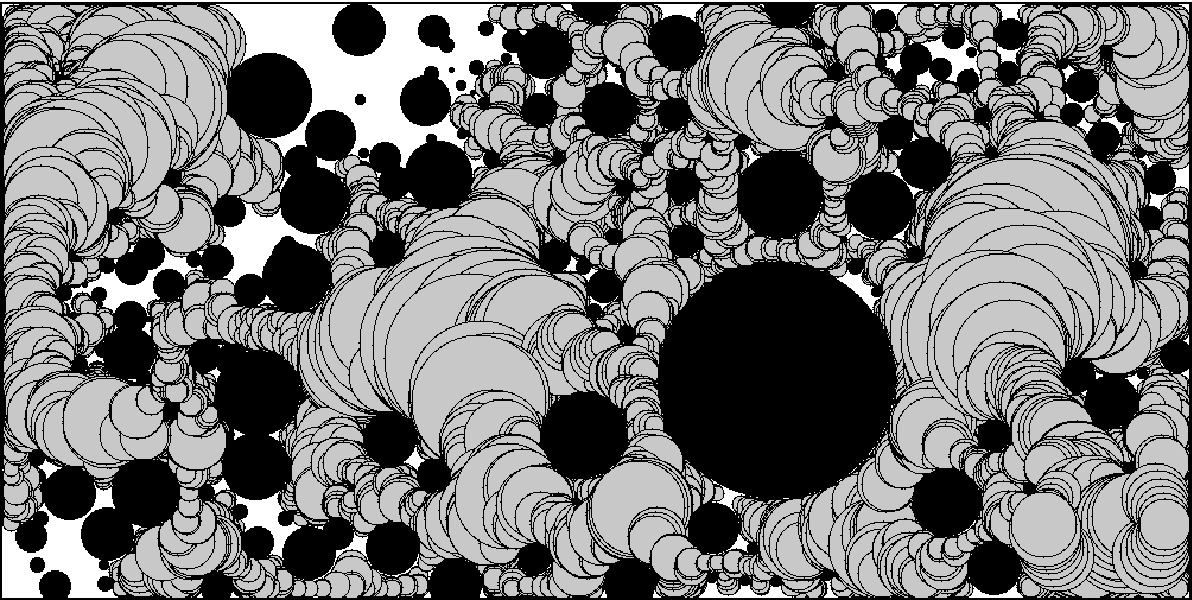}}
  \subfigure[ESD]{\label{fig:sas_normal_1_distribution}\includegraphics[width=0.4\textwidth]{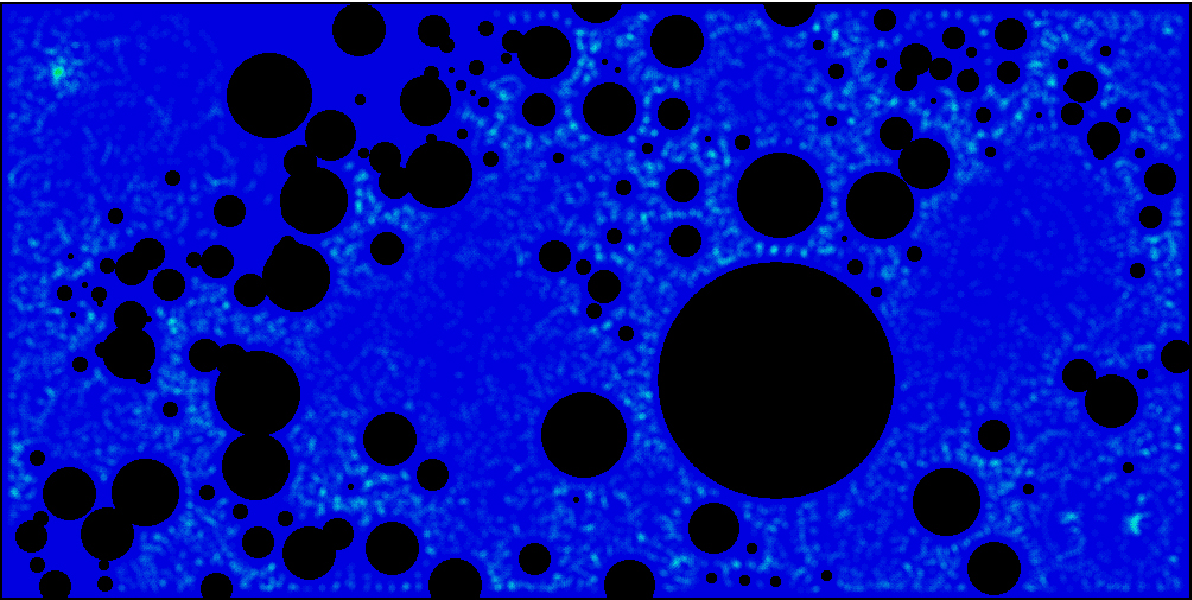}}
  \caption{The result of the SAS by setting $\kappa^g=1$: (a), (b) and (c) show that both the length of the motive primitives and the effective zone adaptively vary more dramatically, when the search approaches the obstacles and the goal. This makes the space very sparsely explored as shown in (d).}
  \label{fig:sas_normal_1}
\end{figure*}

\subsection{A General Environment}\label{sec:general}
Fig. \ref{fig:ccs_normal_path} shows a general environment, in which the circular obstacles are randomly set all over the space. This experiment shows how $\kappa^o$ would affect the behavior of the SAS. The other two coefficients of the SAS are constantly set as $\kappa^g=0.6$ and $\lambda=3$.

Fig. \ref{fig:ccs_normal} shows the result of the Weighted A* by setting $\mu=2.0$. It is observable from Fig. \ref{fig:ccs_normal_distribution} that the local minimum is the area that totally turns red. This means those areas have been fully explored, which seriously trapped the search. The same effect is also observable in Fig. \ref{fig:ccs_normal_tree} and Fig. \ref{fig:ccs_normal_effective}.

Fig. \ref{fig:sas_normal_0.5} shows the test of the SAS with $\kappa^o=0.5$. As in Fig. \ref{fig:sas_normal_0.5_effective}, it is observable that the size of the effective zone varies according to the size of the free space in the local area. The effective zone shrinks when the search reaches narrow areas and is becomes inflated when the search explores open areas.  As shown in Fig. \ref{fig:sas_normal_0.5_distribution}, the areas close to the obstacle mostly turn green, and only very sparse green dots scattered over the open areas. This shows that the SAS focuses on narrow areas, leaving open areas only roughly explored. As in Fig. 11(c), the effective zone also shrinks as the search approaches the goal. This makes the search ignore the states that are relatively far away from the goal, thus, converge to the goal much faster. The SAS adaptively changes the intensiveness of the space exploration, saving substantial computation costs without reducing the precision. Fig. \ref{fig:sas_normal_0.5_path} shows the produced path; it consists of scaled motion primitives of different lengths. Shorter motion primitives are applied as it comes close to the obstacle or the goal to make the search finer. Longer motion primitives are applied to increase the size of the steps to deal with open areas. 

\begin{figure*}[htbp]
  \centering
  \subfigure[Path cost]{\label{fig:general_diagram_path_cost}\includegraphics[width=0.32\textwidth]{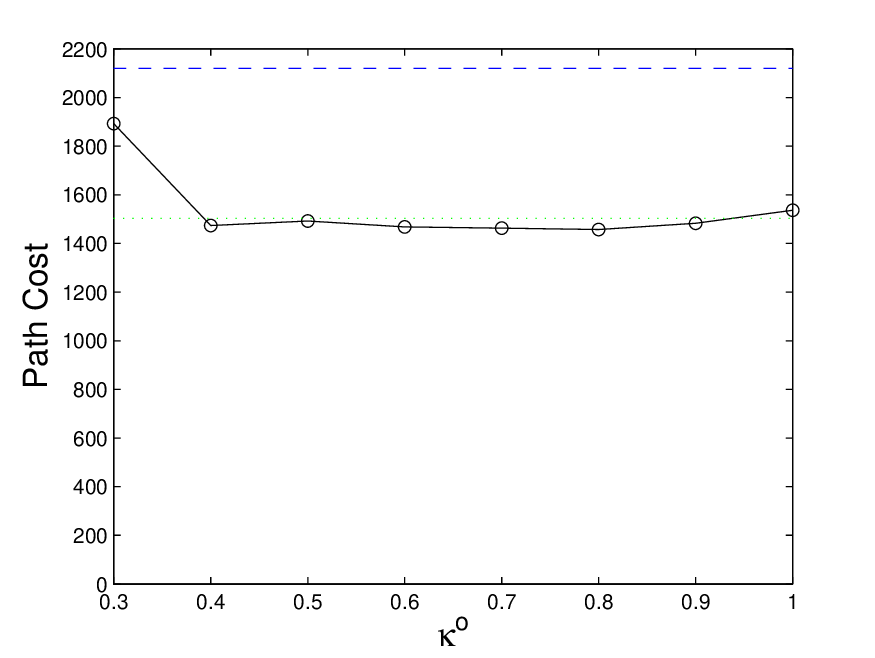}}
  \subfigure[Computation time (ms)]{\label{fig:general_diagram_time_cost}\includegraphics[width=0.32\textwidth]{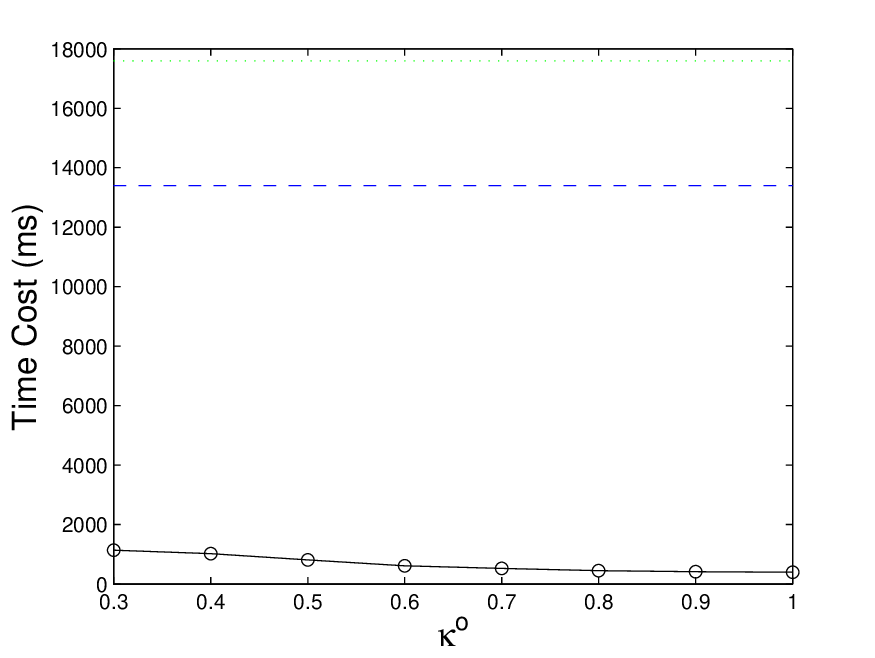}}
  \subfigure[Memory cost]{\label{fig:general_diagram_memory_cost}\includegraphics[width=0.32\textwidth]{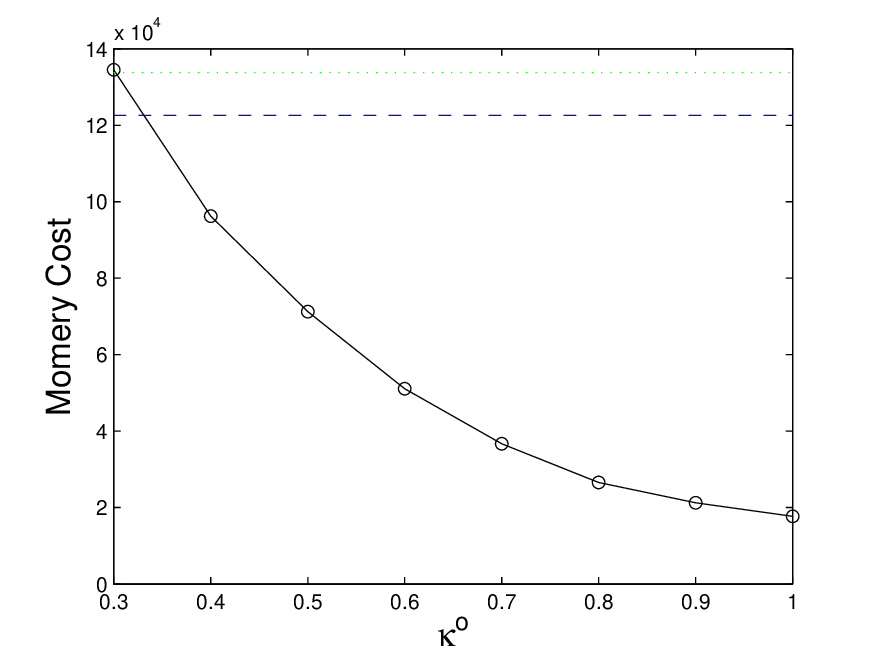}}
  \caption{The Cost variation of the SAS by increasing $\kappa^o$ from $0.3\sim 1$. The green line and the blue line show the result of the Weighted A* by setting $\mu=2.0$ and $\mu=3.0$ respectively.}
  \label{fig:general_diagram}
\end{figure*}

Fig. \ref{fig:sas_normal_1} is the test of increasingly setting $\kappa^o=1$. Compared with the result in Fig. \ref{fig:sas_normal_0.5_distribution}, the space in Fig. \ref{fig:sas_normal_1_distribution} has barely been touched, except the spot around the goal. Based on (\ref{eq:scalar}) and (\ref{eq:effective_radius_final}), the inflation of $\kappa^o$  not only increases the size of the effective zone as in Fig. \ref{fig:sas_normal_1_effective}, but also increases the relative size of the motion primitives as in Fig. \ref{fig:sas_normal_1_tree}. The space has been almost fully explored as shown in Fig. \ref{fig:sas_normal_1_effective} and  Fig. \ref{fig:sas_normal_1_tree}, but only a very limited number of states have been visited as shown in Fig. \ref{fig:sas_normal_1_distribution}. This reviews the reason why the SAS is very efficient.

Fig. \ref{fig:general_diagram} shows the cost variation of the same experiment. The black lines represent the result of the SAS by increasing $\kappa^o$ from 0.3 to 1.0, whereas the result of the Weighted A* is shown as the green line ($\mu=2.0$) and the blue line ($\mu=3.0$). It clearly shows that compared with the Weighted A*, the SAS only requires much less computation time [Fig. \ref{fig:general_diagram_time_cost}] and memory cost [Fig. \ref{fig:general_diagram_memory_cost}], while still maintaining the path cost [Fig. \ref{fig:general_diagram_path_cost}] in an acceptable range. Both the computation time and the memory cost of the SAS dramatically decrease as $\kappa^o$ approaches 1.0. It is also observable that if there are more local minima in the space, the advantage over the Weighted A* can be much more dominant. Increasing $\kappa^o$ would cause more states to be updated in each step, thus speeding up the path planning process. Reducing $\kappa^o$ also reduces the size of each step and, therefore, can increase the flexibility of the search, but this will also increase the computation cost. However, based on our experiment, setting $\kappa^o=1.0$ makes the SAS very efficient and flexible enough to deal with very clustered environments.

\begin{figure*}[htbp]
  \centering
  \subfigure[Path cost]{\label{fig:general_diagram_path_cost_mu}\includegraphics[width=0.32\textwidth]{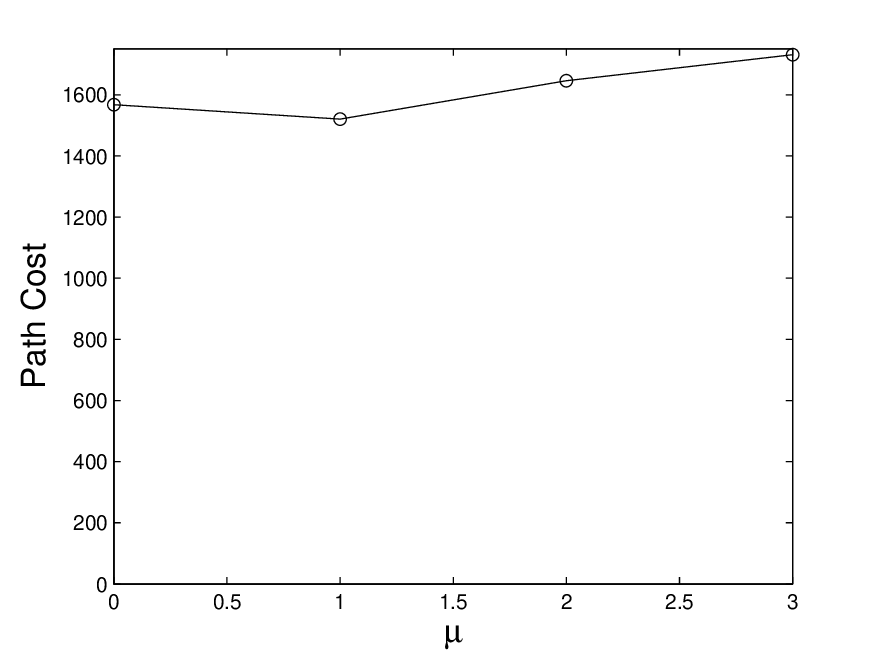}}
  \subfigure[Computation time (ms)]{\label{fig:general_diagram_time_cost_mu}\includegraphics[width=0.32\textwidth]{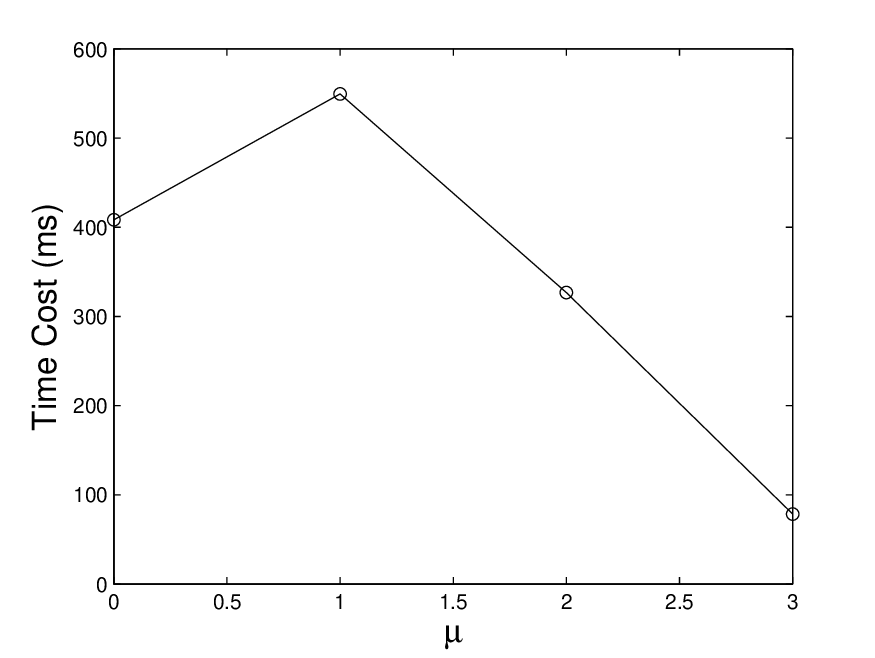}}
  \subfigure[Memory cost]{\label{fig:general_diagram_memory_cost_mu}\includegraphics[width=0.32\textwidth]{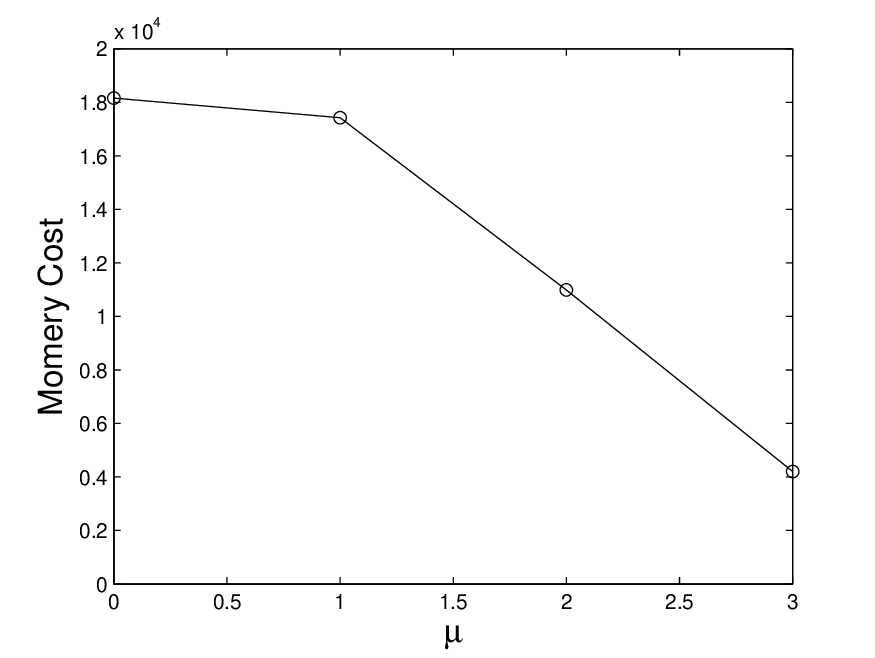}}
  \caption{The Cost variation of the SAS used together with heuristic by increasingly setting the inflator $\mu$ from 0.0 to 3.0. ($\kappa^o= 1.0$ and $\kappa^g=0.6$)} 
  \label{fig:general_diagram_mu}
\end{figure*}

As mentioned earlier, the SAS can be used also together with heuristic. Fig. \ref{fig:general_diagram_mu} shows the result of the SAS applied with heuristic of different inflating coefficient $\mu= 0.0\sim 3.0$ on the same map shown in Fig. \ref{fig:general_diagram}. $\kappa^o$ and $\kappa^g$ are set constant at 1.0 and 0.6 respectively. It is observable that as $\mu$ increases from 0.0 to 3.0, both computation time and memory cost fall substantially while the path cost slightly increases.

\section{Conclusion}
A novel SAS for nonholonomic mobile robot path planning is proposed. Other than universally searching the state-space, the SAS only roughly explores open areas, but preserves high-precision exploration in narrow areas and the areas around the goal.
Since the SAS updates not only the state on the current location but also all states in the effective zone, a large number of areas can be updated immediately in every step. This enables the SAS to get rid of the local minimum very fast. 
Sizes of the motion primitives are also dynamically varied, so the successive steps only slightly get out of the effective zone. This not only promotes the efficiency of the search, but also reduces the number of failed steps that can be blocked by obstacles. Motion primitives can be scaled to different size or curvatures, which make the produced path no longer limited by the predefined set of motion primitives. With a very low computation cost, the SAS is capable of dealing with partially known or unknown environments, in which the robot is required to build or update the map on the fly. In order to increase the clarity of the paper,  we ran the SAS without heuristic which still produced dominate speedup and largely reduced the memory cost. However, the use of the SAS is very flexible, and it can be used together with heuristic or other path planning algorithm.

\appendices

\ifCLASSOPTIONcaptionsoff
  \newpage
\fi

\bibliographystyle{IEEEtran}
\bibliography{sas}

\vfill
\end{document}